\documentclass{article} 
\usepackage{iclr2025_conference,times}

\usepackage{amsmath,amsfonts,bm}

\def\eqref#1{equation~\ref{#1}}

\def\1{\bm{1}}

\DeclareMathAlphabet{\mathsfit}{\encodingdefault}{\sfdefault}{m}{sl}
\SetMathAlphabet{\mathsfit}{bold}{\encodingdefault}{\sfdefault}{bx}{n}

\usepackage{hyperref}
\usepackage{url}
\usepackage{booktabs}
\usepackage{float}   
\usepackage{xcolor}
\usepackage{graphicx}
\usepackage{tabularx}
\usepackage{amsmath}
\usepackage{bm}
\usepackage{enumitem}
\usepackage{siunitx}
\usepackage{comment}

\usepackage[table]{xcolor}    

\definecolor{BestGreen}{HTML}{DFF1D8}
\definecolor{WorstRed}{HTML}{FDE0E0}

\title{Non-identifiability of Explanations from Model Behavior in Deep Networks of Image Authenticity Judgments}

\author{Icaro Re Depaolini, \& Uri Hasson \\
Center for Mind/Brain Sciences\\
The University of Trento\\
Trento, Italy \\
\texttt{\{icaro.redepaolini,uri.hasson\}@unitn.it} \\
}

\usepackage{soul}          

\iclrfinalcopy 
\begin{document}

\maketitle

\begin{abstract}
Deep neural networks can be trained to predict human judgments, but successful prediction does not mean those networks use information similar to that used by humans, nor does it expose that information. To address this issue, prior work has produced explanations, in the form of attribution heatmaps, to identify the cues models use for prediction. This, in turn, relies on the assumption that explanations are robust in the first place. We tested this assumption in the domain of image authenticity, as AI systems can already produce content that appears highly authentic, making it important to understand what drives human assessment of such images. Specifically, we evaluate whether models that predict human ratings produce consistent explanations within and across architectures. We fit lightweight regression heads to multiple frozen, pretrained vision architectures to predict human authenticity ratings and generated attribution maps using Grad-CAM, LIME, and multiscale pixel masking. Several architectures predicted ratings well, reaching about 80\% of the noise ceiling. However, VGG models achieved their performance primarily by tracking image quality rather than authenticity-specific variance, limiting the relevance of the attribution maps they produced. Among the remaining high-performing models, attribution maps were typically stable across random seeds within an architecture, particularly for EfficientNetB3 and Barlow Twins, and this consistency was higher for images judged as more authentic. Most importantly, agreement across architectures was generally weak, even when models achieved similar predictive performance. Given this diversity, we combine information across architectures using ensemble models. We find that ensembles improve behavioral prediction and allow producing image-level attribution through pixel masking. Together, our results show that deep networks can provide good predictions of human behavior, but without providing identifiable explanations for psychologically relevant cues underlying human judgments. More broadly, the results suggest that, at this point, post hoc explanations produced by successful behavioral models should be considered weak evidence for cognitive mechanism.
\end{abstract}

\section{Introduction}
State-of-the-art computer vision models based on deep neural networks are routinely used for core tasks such as classification and segmentation. Their strong performance arises from their ability to learn a large number of effective features during training. These models are considered as psychologically relevant feature extractors, as their representations often align with those of humans \cite[e.g.,][]{tarigopula2023improved, peterson2018evaluating, sucholutsky2023alignment}. In Cognitive Science, Psychology, and Cognitive Neuroscience, these pre-trained models have been used to predict human perception of various image characteristics including image quality \cite[e.g.,][]{zhang2025quality, lu2025deepquality}, authenticity \cite[e.g.,][]{wang2023aigciqa2023}, and naturalness \cite{chen2024Natural}.

These advances are important for both engineering and basic research. For human-facing applications, good performance often benefits from learning a mapping between DNN feature sets and human judgments, because dimensions such as perceived authenticity and perceived quality are defined by how images are experienced by humans. From the perspective of basic research, having models that predict human behavior means that those models may, at least partially, inform us about the mechanisms and representations that underlie that behavior. While successful prediction of human behavior is useful for hypothesizing about shared representations, obtaining \textit{explanations} for why the vision models make those predictions allows a stronger evaluation of this question.  

Here we therefore focus on the \textit{explanatory} potential of computer vision models trained to predict human perceptions of image authenticity. We use the term \textit{explanation} in its technical sense, referring to attribution or saliency heatmaps that indicate which image regions most influenced a model’s prediction. Our goal is to examine whether these explanations are sufficiently stable and consistent to support psychological interpretation. This extends previous work that evaluated model performance \cite[for recent evaluations, see, e.g.,][]{wang2023aigciqa2023, chen2024Natural} but shifts the focus to their potential relevance for understanding human cognition. Our approach is different from image forensics, whose interest is in determining whether an image is objectively real or fake \citep[for review, see][]{wang2024deepfake}. Instead, by modeling human authenticity perception we target subjective processes that underlie human judgments. Prior work has shown that although individuals may perform at chance when distinguishing real from fake faces \citep[at least for some categories, see e.g.,][]{nightingale2022ai}, group‑level judgments are consistent even when people are incorrect. This suggests that people tend to rely on common cues when evaluating authenticity. 

Our aim in the current study also departs from prior work by treating prediction as a means rather than an end. We use machine learning and especially explainable AI to evaluate if models trained on authenticity judgments can, in principle, support explanatory inferences about human behavior. To this end, we study such models and evaluate two successive criteria: first, whether the models track authenticity-specific information rather than a correlated proxy (quality); and second, whether the explanations they produce are sufficiently stable across retrainings within an architecture and sufficiently consistent across architectures to support interpretation.

\section{Related work}
\subsection{Predicting human authenticity ratings}
 There exist many datasets aimed at training computer vision models to predict human perception of image quality \citep[see][]{chen2024Natural}. However, only few datasets also collect ratings of human authenticity perception. AIGCIQA2023~\citep{wang2023aigciqa2023} contains 2400 AI-generated images, each accompanied by human ratings of \textsc{Quality}, \textsc{Authenticity}, as well as the semantic match between the image and the prompt used to create it (\textsc{Correspondence}). For \textsc{Authenticity}, participants rated how real an image appears. The authors report prediction benchmarks for \textsc{Authenticity} on standard backbones (VGG-16/19, ResNet-18/34).  For \textsc{Authenticity}, rank correlations between human ratings and model predictions were comparable across backbones ($r {\sim}0.66$). In later work, two studies reported improved prediction performance ($r {\sim}0.79$) using CLIP-based models whose training includes both textual descriptions of the images and their authenticity rating \citep{tang2024clip, zhou2025cross}.  Pku-aigiqa-4k \citep{yuan2025pku} contains ratings for authenticity, quality and correspondence for 4,000 images. The authors show that multimodal predictors, which provide textual and visual information about the images during authenticity prediction, produce better prediction of human authenticity scores arriving at a correlation of $r{\sim}0.82$.

Related work \citep{chen2024Natural} obtained \textit{naturalness} ratings for AI-generated images, as well as separate ratings for \textit{technical quality} and \textit{rationality}. Naturalness was not explicitly defined, but was implicitly anchored, through an annotation interface, to the presence and extent of unnatural areas in an image. Quality referred to low-level visual properties and distortions (e.g., blur, detail loss, contrast/luminance issues, artifacts). Rationality referred to features related to realism such as whether the arrangement of objects is logical and if the scene could  occur in the real world. Importantly, the study showed a high correlation between quality and rationality ratings ($r=0.81$), suggesting that in AI-generated images, more sensible content is associated with better technical quality. In addition, the authors showed that computer-vision models could be trained to predict naturalness ratings with good accuracy (e.g., $r {\sim}0.83$). 

These findings can be considered from two perspectives: on a purely psychological one, authenticity as a construct might be multifactorial, related to aesthetic, semantic and technical qualities. Co-variance between these factors could make it easier to construct models that predict human authenticity ratings, but at the same time, isolating variance unique to authenticity becomes difficult. Furthermore, two equally effective ML models can arrive at the same performance for different reasons: one, e.g. relying mainly on technical-related factors, and the other on semantic ones. Specifically, prior work \citep{wang2023aigciqa2023} has not evaluated if successful authenticity prediction is related to image-quality. In the current study, one of our aims is to evaluate whether models that successfully predict human authenticity ratings are doing so via quality-related or authenticity-specific variance, using a partial-correlation decomposition.

\subsection{Explainability of models predicting human judgments}

Model performance  does not provide information about which image-features a model relies on for achieving its predictions, which requires the use of explainability methods. Nonetheless, prior work has not provided machine-generated explanations for models that predict human authenticity perception, nor has it tried to define independent criteria for determining when explanations are useful. Most generally, in the area of explainable AI, \textit{explanations} refer to methods that attribute importance to input regions by highlighting those regions whose variation is associated with changes in the model’s output. These methods therefore provide indications for what information the model uses to generate its output. In image classification, explanations indicate image sections whose content is important for correct classification ~\citep[for review, see][]{mohamed2022review}. For example, gradient-based methods quantify how changes in internal unit activations impact the model's prediction (i.e., they identify features that strongly support correct predictions). In contrast, perturbation-based methods \citep[][]{zeiler2014visualizing, palazzo2020decoding} mask or blur parts of the original image and measure how this changes the output. In the context of predicting human image authenticity, these methods can indicate which image sections contribute positively or negatively to the model’s estimate of authenticity. We note that not all computational models can produce explanations. For example, large-language models (LLMs) and multi-modal LLMs have been used for predicting image authenticity~\citep[for recent review, see][]{zhang2025quality}, but in these cases, interpretability is limited by their complex structure.

Several prior studies have applied explainability approaches to models trained to predict human judgments (e.g., visualizing cues for aesthetics ~\citep[e.g.,][]{tong2022interpretable}, or memorability~\citep[e.g.,][]{khosla2015understanding}. These explanations, by design, provide insight into the model’s decisions, though it is less clear if they also provide insights into the information used for the human judgment. 

This apparent tension between i) the availability of an explanation for the model's prediction of human behavior and ii) the model's explanatory relevance for human behavior is not contradictory: it is sufficient to consider that different models with similar performance can still produce divergent explanations. Furthermore, agreement in predictions or in internal representations does not guarantee agreement in attribution-based explanations. This situation where diverse, similarly-performing ML models still produce different explanations has been referred to as the \textit{Rashomon effect},~\citep[e.g.,][]{muller2023empirical, gwinner2024comparing}, and has also been documented for heatmap-based explanation methods~\citep[e.g.,][]{li2023cross, watson2022agree}.  It follows that if a Rashomon effect is documented for models that successfully predict human authenticity ratings, this would suggest that substantial caution should be exercised when using these models to draw inferences about human psychology.

\section{Overview and aims}
\begin{figure}[t]
    \centering
    \includegraphics[width=1\linewidth]{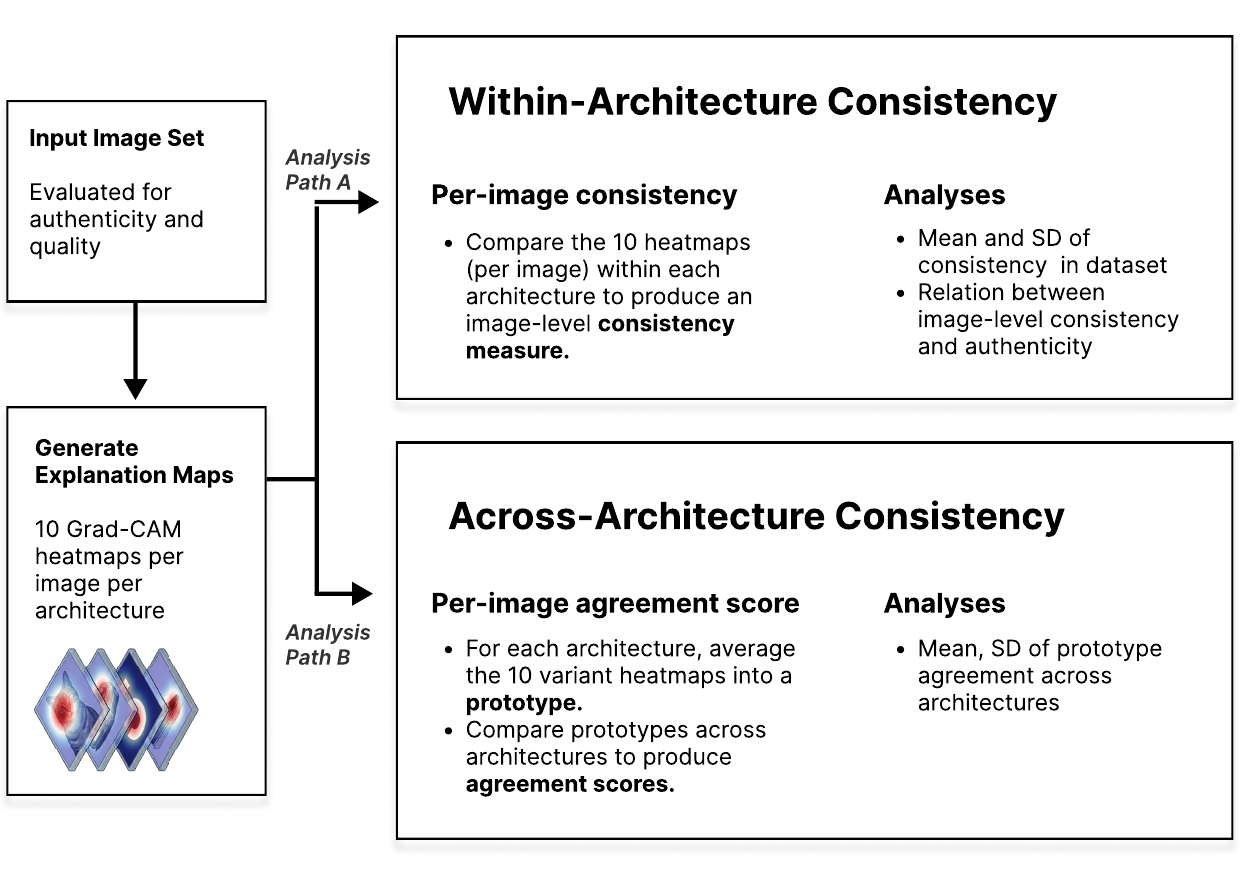}
    \caption{\textbf{Explanation consistency analysis: within- and across-architecture.} For each image and each architecture, 10 explainability maps were generated from 10 independently trained model variants. \textit{Within-architecture} consistency was computed as the mean pairwise similarity among the 10 heatmaps per image. \textit{Across-architecture} consistency was computed by first averaging the 10 maps per architecture into a single prototype, then correlating prototypes across all architecture pairs, producing one agreement score per image per pair.}
    \label{fig:approach}
\end{figure}

As summarized above, prediction accuracy cannot be treated as a sufficient criterion for psychological relevance, and the consistency of model-generated explanations has not been studied in the domain of human authenticity prediction. Given these considerations, we determine the relevance of deep vision models to  human  perception of image-authenticity using a two-pronged approach. In the first step, we evaluate the predictive performance of several pre-trained computer-vision architectures by freezing the pre-trained backbones and adapting them to predict human \textsc{authenticity} ratings provided by participants. Here we follow the approach of \citet{wang2023aigciqa2023}, which fits a simple adapter to frozen, pre-trained backbones. This constrains the information to each backbone's pretrained features. Freezing the backbone is essential given the relatively small training sets and the large number of parameters that would be involved in retraining the backbone itself.

In a second step (see Figure \ref{fig:approach}), we evaluate the consistency of model explanations. Here we evaluate two questions: 1) to what extent does each architecture produce consistent explanations when trained repeatedly from different random seeds, and 2) to what extent do different architectures produce similar explanations.  We ask the first question because if a given single architecture produces inconsistent explanations (when retrained from different seeds), then it is difficult to treat any of those explanations as potentially relevant to human perceptual representation. We ask the second question because even if two architectures are internally consistent, they may still provide different explanations, which would make it difficult to judge which architecture, if any, provides a better analog of human processing.   

We have four main aims:
\begin{itemize}[leftmargin=*, labelsep=0.5em]
    \item Evaluate the predictive effectiveness of different architectures in predicting human authenticity judgments and whether their predictions reflect authenticity-specific versus quality-shared variance.
    \item Determine whether different architectures provide internally-consistent (within-architecture) explanations across random initializations.
    \item Determine whether different architectures provide consistent across-architecture explanations for authenticity prediction.
    \item Test whether ensemble models can aggregate information over architectures to improve prediction accuracy while still providing explanations when using appropriate methods.
\end{itemize}

Anticipating the results, we find that some architectures learn to predict human authenticity judgments by learning features related to quality rather than authenticity proper. Of the remaining architectures, some showed low within-architecture explanation consistency across random seeds, and cross-architecture consistency was generally weak. Ensemble predictors provided the strongest alignment with human judgments, and produced attribution maps at the image level.

\section{Methods}
We present three experiments that share the same backbone and regression-head  architecture but differ in how data partitions are used. In Experiment~1, we train models using a standard train/validation/test design to evaluate prediction of human authenticity judgments and its relation to quality information. Experiment~2 uses the same trained models but  changes the role of the test partition so that it is used as set for which explanations are optimized. Experiment~3 extends the design to ensemble learning, but uses a larger test partition to support meta-learning.  Table~\ref{tab:partition_roles} summarizes the partition roles across settings; the subsections below describe each component in detail.

\begin{table}[t]
\centering
\small
\begin{tabular}{lcccp{6.6cm}}
\toprule
\textbf{Setting} & \textbf{Base fit} & \textbf{Model sel.} & \textbf{Final eval.} & \textbf{Implementation} \\
\midrule
Exp1 & 70\% & 20\% & 10\% &
Train/validation used for early stopping; the selected model is evaluated on a held-out test set. \\

Exp2 & 70\% & 20\% & 10\% &
Train/validation used for early stopping; pruning is optimized on the test partition (used here as an explanation-optimization set). \\

Exp3\\ Bagging & 70\% & 10\% & 20\% &
A fixed 20\% test split is held out. 60 variants are trained on different train/validation splits of the remaining 80\%; ensemble prediction obtained by averaging test predictions across models and evaluated on the held-out test set. \\

Exp3\\ Stacking & 70\% & 10\% & see Imp. &
A fixed 20\% partition is used for stacking. It is split internally for 5-fold CV (meta-train/meta-holdout); reported performance is averaged over held-out folds. \\
\bottomrule
\end{tabular}
\caption{\textbf{Dataset partition roles across experimental settings.} Percentages refer to the full dataset ($N=1367$). For Exp3 (Bagging and Stacking), percentages describe the split applied to each individual base model.}
\label{tab:partition_roles}
\end{table}

\subsection{The AIGCIQA2023 dataset}
We used the AIGCIQA2023 dataset, which evaluates human visual preferences using three different ratings: \textsc{Quality}, \textsc{Authenticity}, and \textsc{Correspondence}. The database consists of 2,400 AI-generated images (AIGI) produced by six different generative models, that were prompted to produce images belonging to a variety of pre-defined categories. For each image, participants rated Quality, Authenticity and Correspondence between the text prompt and the produced image. Our interest is in authenticity, which was operationally defined as whether the image seemed natural or AI generated.  Human judgments were collected as individual raw ratings on a scale from 1 to 5. Mean Opinion Scores ($\mathrm{MOS}$) were then calculated from these raw ratings. For each participant, scores were first Z-scaled, and then rescaled to a 0–100 rating scale. 

The dataset's images were generated from prompts that refer to particular categories, combined with different types  of  'challenges' (e.g., asking for simple vs. detailed images for the category). Some combinations of categories and challenges produced images that are, \emph{by design}, non-realistic. For this reason, before any modeling, we defined an exclusion rule based solely on prompt metadata and removed those images (Removed $n = 1033$; Kept $n = 1367$) and do not consider them further (sample non-natural categories include \textit{illustrations}, \textit{arts}, and \textit{imagination}; a complete list of images removed is available on GitHub). To validate the removal, we computed for each participant the mean authenticity scores of the kept and removed images and compared those via paired Student's T-test ($n=25$). We found that the removed images had lower authenticity ratings (Means = 2.18 vs.\ 2.29 $t(24) = 5.3, p < 0.001$, Cohen's $d=1.06$); see Appendix Figure \ref{fig:AppRemovedImages}. Similar differences were found for \textsc{quality} (Means = 2.26 vs.\ 2.40 $t(24) = 3.69, p < 0.01$, Cohen's $d=0.74$) and \textsc{correspondence} (Means = 2.20 vs.\ 2.35 $t(24) = 7.58, p < 0.001$, Cohen's $d=1.5$). 

\subsection{Human reliability and correlation ceilings}

\paragraph{Human reliability.}~In classical test theory, the correlation between an observed test score and the unobserved true score equals the square root of the test’s reliability \citep{NunnallyBernstein1994}. We therefore estimated a noise ceiling from the Spearman-Brown-corrected split-half reliability of the mean ratings.\footnote{We thank the authors of the AIGCIQA2023 dataset for sharing with us the raw single-participant data required for computing these reliability estimates.} Although our primary focus is \textit{authenticity}, we assessed split-half reliability for all three behavioral measures (\textsc{quality}, \textsc{authenticity}, and \textsc{correspondence}) to characterize the dataset as a whole. For each measure, we constructed an item-by-participant matrix and repeatedly (20 times) split participants randomly into two groups (12 and 13 participants). For each split, we computed item-wise mean ratings within each group, producing two item-level vectors, and computed the Pearson correlation. Correlations were Fisher-$z$ transformed, averaged across splits, and the resulting mean correlation was converted back to $r$ and corrected with the Spearman--Brown formula to estimate reliability. The human noise ceiling was then taken as the square root of this reliability.

\paragraph{Model prediction reliability.}
To quantify the stability of each architecture’s predictions, we estimated model reliability from agreement across independently trained instances of the same model. Specifically, for each architecture we trained 10 variants differing only in random initialization and the objects' training order (random seed). For each variant, we stored the predicted \textit{authenticity} score for every test image. This produced a prediction vector (with $N=137$ test images) per variant. We then computed the Pearson correlation between prediction vectors for all pairs of variants and defined model reliability as the mean pairwise correlation across the $\binom{10}{2}=45$ pairs. Finally, to obtain an adjusted ceiling on the observable model--human correlation, we combined the human noise ceiling with the model reliability under the assumption that both are noisy measurements of an underlying latent image-wise score, $r_{\max} \approx \sqrt{r_{\text{human}}\cdot r_{\text{model}}}$.

\subsection{Backbone architectures}
As backbone models for authenticity prediction we used the following supervised models: VGG-16 and VGG-19 \citep{simonyan_very_2015}, ResNet152 \citep{he_deep_2015}, DenseNet161 \citep{huang_densely_2018}, and EfficientNetB3 \citep{tan_efficientnet_2020}. In addition, we used a backbone trained using a Barlow Twins architecture \citep{zbontar_barlow_2021}, which is a self-supervised model whose objective is to maximize the similarity between the representations of two augmented views of the same image while minimizing the redundancy between information coded by different model units.
 
\subsection{Training setup and objective}
\label{sec:trainsetup}
We used transfer-learning where the pretrained backbones were frozen and used as feature extractors. A regression head $h_{\theta}$ (two hidden layers with ReLU and dropout $p{=}0.5$, followed by a linear scalar output) mapped the backbone embeddings. Images were resized to $256{\times}256$, center-cropped to $224{\times}224$, and normalized with ImageNet statistics. 

To create 10 variants of each model, we used a repeated random splitting procedure (Monte-Carlo cross-validation). Specifically, we generated $S=10$ independent stratified splits of the $N=1367$ images into 70/20/10 train/validation/test sets (no image appears in more than one set within a split). We report average performance across splits. We used the Adam optimizer (learning rate $10^{-3}$), batch size 64, training for up to 250 epochs with early stopping (patience = 15 epochs), at which point the last-stored best model was used. The prediction target was the human Mean Opinion Score ($\mathrm{MOS}$) of \textsc{authenticity} on a $[0,100]$ scale. For each batch we minimize mean squared error ($\mathrm{MSE}$) of the prediction. 

\paragraph{Diagnostic: authenticity-specific versus quality-shared variance.}
Because \textsc{authenticity} and \textsc{quality} ratings are strongly correlated in the dataset (see Section~\ref{sec:results_psychometric}), a model that predicts \textsc{authenticity} may do so by tracking \textsc{quality}-related image dimensions rather than unique variance contributed by authenticity. To assess this, we evaluated the degree to which each architecture's predicted authenticity score ($\hat{A}$) correlated with the observed \textsc{quality} ratings ($Q$), and also computed the partial correlation $r(A,\hat{A}\mid Q)$. This latter term reflects the residual association between true and predicted \textsc{authenticity} after removing the linear contribution of \textsc{quality} from both variables. A near-zero partial correlation indicates that the model captures little authenticity variance beyond what is shared with quality.  

\subsection{Channel pruning for explanation-set readout}
\label{subsec:sbs-pruning}

\paragraph{Rationale for explanation-set pruning.}
The regression head was trained using the test/val splits to predict authenticity across all training images, and the head's weights define a useful readout of the backbone's feature space. When generating Grad-CAM explanations for a specific held-out set of images (the \textit{explanation set}, here coinciding with the test split), we wanted the attribution
map to reflect those channels feeding into the regression head which are most relevant given a particular head and particular run. Other, less relevant channels may contribute gradient noise without contributing to the prediction, and are therefore less important for identifying the attribution signal. We therefore implement unit-pruning via Sequential Backward Selection (SBS; see below) to identify, for each trained regression head, the minimal subset of channels in the target layer (the last convolutional layer of the frozen model) that is sufficient to predict authenticity on the explanation set. SBS identifies a specific read-out for the explanation set \textit{without modifying} the head's weights or recombining its feature directions. In other words, pruning removes less important channels from the pre-fixed readout while leaving all other elements fixed. This stands in contrast to fine-tuning the regression head, which would reweight and mix features so that the resulting directions themselves are optimized for the explanation set. We note that predictive performance after explanation-set pruning cannot be interpreted as generalization, as characteristics of the  explanation set itself are used to optimize the pruning mask. We use the pruning solution here only for the explanation pipeline.

\paragraph{Sequential backward selection (SBS) for pruning.}
\label{sec:sbs}
Pruning was implemented using a greedy \emph{Sequential Backward Selection} (SBS) wrapper procedure \citep{guyon2003introduction}. At iteration \(t\), let \(M^{(t)}\) be the current model and consider each remaining feature map \(k\) in the target layer. We ablate \(k\) by zeroing its convolutional kernel (and bias) and re-evaluate test-set error; the importance of \(k\) is the $\mathrm{RMSE}$ change \(\Delta_k \equiv \mathrm{RMSE}_t(k)-\mathrm{RMSE}_t\), where \(\mathrm{RMSE}_t \equiv \mathrm{RMSE}(M^{(t)})\) and \(\mathrm{RMSE}_t(k) \equiv \mathrm{RMSE}(M^{(t)}\setminus k)\). We select \(k^{*}=\arg\min_k \mathrm{RMSE}_t(k)\) that produces the lowest post-ablation RMSE. If \(\Delta_{k^{*}}<0\) (i.e., ablation improves performance), we remove \(k^{*}\) and set \(M^{(t+1)}=M^{(t)}\setminus k^{*}\); otherwise the process stops. No fine-tuning is performed within an iteration. Iterations continue until no further decrease in $\mathrm{RMSE}$ is reached. Pruning was applied to a single target convolutional layer in each architecture, which was also the layer used for Grad-CAM attribution (see Appendix Table~\ref{tab:pruning_target_layers}).

\paragraph{Implications for interpreting consistency.}
Because the backbone architectures were frozen, their feature maps were fixed across all random seeds. Seed-to-seed variability in the Grad-CAM maps was therefore determined only  by the regression-head initialization weights and the assignment of images to training and validation sets across Monte Carlo splits. Because pruning does not change the backbone's representation, within-architecture consistency in the explanation maps reflects whether different random initializations of the regression head learn similar readout directions from the common feature space of the backbone.

\subsection{Explanation methods: Grad-CAM, MPM, and LIME}
We use three explainability methods. The first is Grad-CAM \citep{selvaraju_grad-cam_2020}, which is an attribution-based method that requires access to model components and gradients. Grad-CAM has been shown to pass basic sanity checks for attribution methods, including sensitivity to model-parameter randomization and data randomization \citep{adebayo_sanity_2020}. We additionally use two perturbation-based methods: Multi-scale Pixel Mapping \citep[MPM;][]{palazzo2020decoding}, and Local Interpretable Model-agnostic Explanations \citep[LIME;][]{ribeiro2016LIME}. These methods quantify the effect of masking local pixel patches (MPM) or pre-defined tesselation (LIME). The  intuition behind perturbation methods is that if removing or masking a part of the input significantly affects the prediction, that part was necessary for the model's decision.  

\subsubsection{Grad-CAM}
We adapt Grad-CAM \cite{selvaraju_grad-cam_2020} to a regression setting using the scalar regression output instead of the class logit. For each feature map in the deepest convolutional layer (see Table \ref{tab:pruning_target_layers} Appendix for layer specification), we compute gradients of the scalar regression output with respect to this layer and obtain channel weights by global average pooling. The Grad-CAM map for this layer is the weighted sum of feature maps. We then upsample this image to the input resolution via interpolation. The resulting heatmap highlights regions that influence the regression output, where positive values (red) indicate a positive shift in the prediction value. Negative values (blue) conversely indicate a negative shift.

\subsubsection{Multi-scale pixel masking (MPM)} MPM is based on the principle of identifying important image patches by evaluating the impact of occlusion on loss magnitude \citep{zeiler2014visualizing}, and here we specifically follow the procedure presented in \citet{palazzo2020decoding} for multi-scale masks, but apply it to evaluating the mask's impact on the regression output. For each pixel $(x,y)$, we applied a binary  mask $m_s(x,y)$ that zeroed out a square patch of size $s\times s$ centered at that pixel (1 elsewhere), producing a masked image $i_{s,x,y}=m_s(x,y)\odot i$. To capture fine and coarse spatial components we applied masking at three scales $\mathcal{S}=\{3,17,65\}$ with stride~1. The importance of each pixel was then defined as the mean drop in predicted score across scales, $S(x,y)=\tfrac{1}{|\mathcal{S}|}\sum_{s\in\mathcal{S}}\Delta_s(x,y)$, 
where $\Delta_s(x,y)=R(i)-R(i_{s,x,y})$. For visualization, importance values were min--max normalized to $[-1,1]$ . Consequently, warm colors (red) indicate regions that positively contribute to the predicted score (masking them causes a drop), whereas cold colors (blue) indicate regions that suppress the score.

Because producing MPM attributions requires evaluating the impact of masking at every pixel location across all model variants, the computational cost is prohibitive for within-architecture consistency analysis (around 2 hours, per image, across 60 variants). We therefore restrict MPM to two analyses: across-architecture agreement, for which we computed attributions for all images using a single variant per architecture, and ensemble attribution, for which we computed attributions across all 60 variants for a small sample of images.

\subsubsection{LIME explanations}
\label{sec:lime}

Grad-CAM and MPM, while often used as explanation methods, have several and distinct weaknesses. Grad-CAM computes gradients in relation to an intermediate convolutional layer. The heatmaps it produces can be excessively smooth because they reflect upsampled information from low-resolution feature maps. Furthermore, it can identify certain channels as important simply because they carry \textit{generally} useful information. In other words, these channels would be important to prediction of any image-related content. In contrast, MPM does not depend on gradients, but its perturbation operator is agnostic to image content (it masks pixels according to a predefined scheme), and its application can be computationally demanding.

For these reasons, we used LIME as a third explanatory method. Like MPM, LIME evaluates the effect of perturbing (masking) image-level regions without requiring access to internal model components or gradients. However, it has the advantage of defining perturbations not over arbitrary pixels, but over an explicit, pre-computed segmentation of the image, where each segment is treated as a feature in a regression model. Here we define segments as SLIC superpixels \citep{achanta2012slic}: based on local color and texture structure, SLIC tessellates the image into compact regions. LIME then operates at the level of the single image by fitting a simple linear surrogate model locally around that image, in order to assign an importance score to each tessellated region (for an example of a tessellated input image and the resulting importance map, see Appendix Figure~\ref{fig:supmatLIME}).

The goal of LIME is to approximate the predictions produced by the original DNN encoder by learning an interpretable regression model that reproduces the DNN's outputs in a local neighborhood around a given input image. While LIME is most often used to explain classification decisions, it can also be applied in regression settings, and here we follow the general logic of Image-Reg-LIME \citep{vinogradova2023local}. For each image $x$ for which an explanation is produced, we first use SLIC to segment the image into $K$ superpixels ($K$ is determined by  SLIC per image, with a user-specified  upper limit, here 150). A perturbation is then defined as a binary vector $z \in \{0,1\}^{K}$, where $z_k=1$ indicates that superpixel $k$ is kept intact and $z_k=0$ indicates that it is replaced by a baseline value (we use the image mean RGB). We generate $M=1200$ perturbations per image; for each perturbation, each superpixel is kept independently with probability $p=0.7$. We pass each perturbed image $x(z)$ through the trained DNN authenticity predictor to obtain a scalar prediction $\hat{y}(z)=f(x(z))$. This produces $M$ predictions, each paired with a perturbation vector.

We then apply LIME to fit a regression surrogate whose objective is to predict the DNN outputs using \emph{only} the perturbation vectors $z$. It learns a mapping $z \mapsto \hat{y}(z)$ (not an image-to-score model), such that the surrogate reproduces the DNN predictions as closely as possible over the sampled perturbations; good agreement indicates that the surrogate captures the DNN's local behavior in the neighborhood of $x$. To do so, we fit a ridge-regularized linear surrogate $g(z)=\beta_0+z^\top\beta$,  with the goal that $g(z)$ approximates the DNN predictions $\hat{y}(z)$ as closely as possible. Specifically, we estimate $(\beta_0,\beta)$ by minimizing a locality-weighted squared error between surrogate outputs and model predictions over the sampled perturbations \citep{ribeiro2016LIME}. In addition, we also use locality weights $\pi(z)$, which decay with the distance of the perturbed vector from the non-perturbed form (all segments are kept, i.e., $z=1$). In this way, perturbations that preserve more superpixels receive higher weight. For interpretability, we fit the surrogate to centered targets, $\hat{y}(z)-\hat{y}(\mathbf{1})$.  This produces interpretable weights where positive betas (red/warm colors) indicate superpixels whose presence increases predicted authenticity; that is, when that region is masked by the baseline color, the predicted authenticity decreases relative to the baseline unperturbed image. Negative betas (blue/cold colors) indicate superpixels whose presence decreases predicted authenticity (masking increases authenticity prediction).

An important difference between LIME and the other two explanation methods is that LIME is a two-stage procedure where the original predictor is first approximated with the surrogate model, and then the parameters (betas) of the surrogate are interpreted. In MPM and Grad-CAM, attribution of importance is directly related to the original predictor, but in LIME, the explanations are only meaningful if LIME effectively approximates the original model's prediction (i.e.,$\hat{y}(z) \approx g(z)$). If the approximation is poor, the resulting coefficients $\beta$ are not a reliable summary of the predictor's behavior and cannot be treated as explanations. For this reason, we quantify per-image surrogate fidelity by computing the locality-weighted fit of $g$ to the DNN outputs, and we interpret LIME maps considering this fidelity measure.

\subsection{Quantifying explanation consistency}

\paragraph{Within-architecture consistency metric.}

For each architecture, we quantified the consistency of its pruned variants in producing explainability heatmaps. For each test-set image 
\(i \in \{1,\dots,138\}\), we generated heatmaps from the 10 pruned variants of
that architecture, computed all 45 unique pairwise correlations and averaged those to produce a per-image average correlation $\bar{r}$. We additional defined consistency as the intersection-over-union (IoU) concordance between any two images after thresholding their attribution maps at 
$\delta \in \{\text{top }5\%,\,15\%,\,25\%\}$.

\paragraph{Across-architecture consistency metric.}

To quantify the similarity of heatmap explanations across
architectures, we first computed a prototypical heatmap for each image and architecture. For every image \(i \in \{1,\dots,138\}\), we averaged the 10 variant heatmaps of a given architecture \(a\) to obtain a single prototype: $P_i^{(a)}$. Then, for each pair of architectures \((a,b)\), we computed the correlation between their prototype heatmaps for each image: $r_i^{(a,b)} \;=\; r(P_i^{(a)},\, P_i^{(b)})$. For each pair of architectures we averaged these correlations across all 138 images to obtain a single across-architecture consistency scores.

\paragraph{The relation between explanation consistency,  prediction accuracy, and authenticity rating.}~We evaluated if higher within-architecture explanation consistency is associated with improved model accuracy or better prediction of human judgments. This analysis was performed separately for the test-set images and the training-set images.  For each image, we computed two values: 1) the consistency across the explanations within each architecture as described above using Intersection-over-union and 2) its average prediction error across the 10 models computed as the average of the absolute deviation (Mean Absolute Error; MAE). We then computed the correlation between these two statistics. We expected that images with higher explanation consistency would be predicted with greater precision, which would produce a negative correlation between consistency and prediction error. 

We additionally evaluated whether explanation consistency correlates with the magnitude of the human authenticity rating. This was an exploratory analysis, as the relation could be either positive or negative. On the one hand, low-authenticity images could have particular visual characteristics (e.g., inconsistent shading or particular non-realistic textures) that would be consistently picked up by models trained from different seeds when using Grad-CAM, in this case producing a negative correlation between consistency and authenticity score.  An alternative is that human judgments of authenticity reflect identification of meaningful realistic elements also picked up by the model, which would produce a positive correlation between explanation consistency and authenticity.

\subsection{Ensemble Methods}
We produced ensembles using a standard approach based on the performance of single-models on out-of-sample test set.  As input to the ensembles we used 60 model variants (10 per architecture) produced as explained in Section \ref{sec:trainsetup}. When creating models for ensemble learning we further used a model-pruning solution (using sequential backward selection; Section \ref{sec:sbs}), where the pruning solution was learned on the \textit{validation} set of each model. We applied this fine tuning because it produced slightly improved generalization performance for out-of-sample test-set predictions. Importantly, for all ensembles, the test-set was isolated during the model selection and pruning stages so that it remained independent for the final evaluation (see Table \ref{tab:partition_roles} for partitions and their roles). 

\paragraph{Bagging (Bootstrap Aggregating)} 
We trained an ensemble by aggregating the predictions of all 60 pruned model variants mentioned above. Given that we trained 10 variants per model, the variations required for bagging was achieved through the unique, randomly seeded train/validation splits as well as the random initialization of regression heads for each of the 60 learners. At inference, predictions on the held-out global test set were aggregated via simple averaging of the variants' predictions. We report ensemble performance using $\mathrm{RMSE}$ and human-alignment metrics ($\mathrm{PLCC}$, $\mathrm{SRCC}$) on the held-out test set.

\paragraph{Stacking (Level-1 Meta-Learning)} 
Here we used the 60 pruned base-learners for stacking-based meta-learning. The global test set served as the basis for $K=5$-fold cross-validation to train and evaluate the meta-learner. Specifically, the test images were partitioned into five folds; for each fold $k \in \{1, \dots, K\}$, one subset (20\% of the test set) was the held out set, and the other four folds (80\% of the test set) were used to train a level-1 meta-learner. 

The meta-learner, implemented as a linear regression layer, was trained on the predictions of the pruned models (acting as input features) against ground-truth labels. This process was repeated across all folds so that every image in the test set was used exactly once for evaluation and $K-1$ times for training. The meta-learner optimizes the prediction:
\begin{equation}
\hat{y}_{\text{stack}}(x) = \mathbf{w}^{\top}\mathbf{p}(x) + b
\end{equation}
where $\mathbf{p}(x) \in \mathbb{R}^{60}$ are the pruned base-model predictions and $(\mathbf{w}, b)$ are the learned parameters. After completing all folds, the out-of-fold (OOF) predictions were combined to evaluate the final performance using $\mathrm{RMSE}$, $\mathrm{PLCC}$, and $\mathrm{SRCC}$.

\section{Results}

\subsection{Psychometric properties of authenticity ratings}
\label{sec:results_psychometric}

Split-half reliability for \textsc{quality}, \textsc{authenticity}, and \textsc{correspondence} were 0.61, 0.50, 0.53. After Spearman–Brown correction, the corrected values were 
0.76, 0.67 and 0.69 respectively, with estimated noise-ceilings of 0.87, 0.82, and 0.83 respectively. This means that the maximum Pearson correlation that a  model could produce with the observed 25-participant mean vector is $\approx 0.82$.

\begin{figure}[t]
    \centering
    \includegraphics[width=\linewidth]{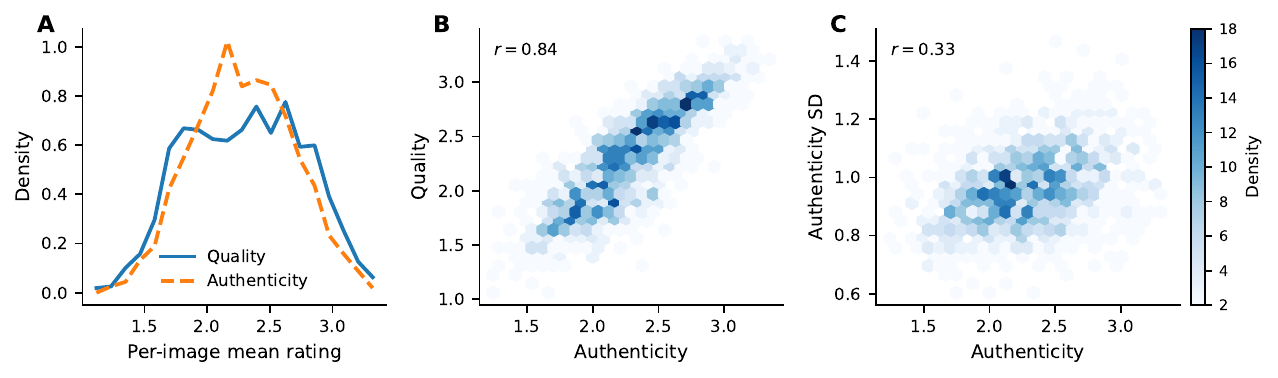}
    \caption{\textbf{Distribution and relationships of human authenticity and quality ratings.} (A) Per-image mean \textsc{quality} and \textsc{authenticity} ratings. \textsc{Quality} showed a bimodal distribution while \textsc{authenticity} did not. (B) Positive correlation between per-image mean \textsc{authenticity} and \textsc{quality} ratings ($r=0.84$). (C) Relationship between mean \textsc{authenticity} and inter-rater standard deviation: images rated as less authentic showed greater rater agreement.}
    \label{fig:auth_quality}
\end{figure}

\textsc{quality} and \textsc{authenticity} were quite strongly correlated (Pearson's $r=0.84$; (Figure \ref{fig:auth_quality}B), but the two statistics had different distributions (Figure \ref{fig:auth_quality}A). A Gaussian mixture bimodality analysis suggested that the \textsc{quality} distribution was  bimodal ($\Delta BIC=65.61$; positive $\Delta BIC$ supports the bimodal model), whereas the \textsc{authenticity} distribution was not ($\Delta BIC < 0$). Given our interest in \textsc{authenticity} we also evaluated the relation between an image's mean authenticity scores (across the 25 participants providing data) and the standard deviation of those scores. For this analysis we first z-scaled each participant's ratings to control for use-of-scale effects.  We found a moderate positive correlation (Figure \ref{fig:auth_quality}C), indicating that participants presented greater agreement for images perceived as less authentic (Pearson's $r = 0.33$). 

\subsection{Model prediction of human authenticity ratings}

\subsubsection{Predictive performance across architectures}
Table~\ref{tab:baseline_and_ceiling} summarizes prediction error ($\mathrm{RMSE}$) for \textsc{authenticity} across architectures, as well as the correlation between model predictions and human ratings ($\mathrm{SRCC}$, $\mathrm{PLCC}$). Note that models were trained to  minimize $\mathrm{MSE}$ alone. Overall, Barlow Twins achieved the best accuracy (lowest $\mathrm{RMSE}$). DenseNet161  presented slightly higher $\mathrm{RMSE}$, but similar $\mathrm{PLCC}$ and $\mathrm{SRCC}$. Interestingly, while  Barlow Twins uses ResNet backbone as its feature extractor, it outperformed ResNet152 both on $\mathrm{RMSE}$ and PLCC/SRCC. Furthermore, not only did ResNet152 produce the weakest predictive performance, its predictions were also highly consistent across random initializations, as reflected in high model reliability (mean Pearson $r=0.91$). That is, ResNet152 was  consistently inaccurate. 

While Table~\ref{tab:baseline_and_ceiling} reports average $\mathrm{RMSE}$ and correlation values over ten variants per model, it is important to note that prediction of human ratings from any \textit{single} model variant, a typical usage case, would be constrained by the reliability of the human data (here $\sim 0.67$) as well as the reliability of the model’s own predictions across random seeds. For a single trained instance of a given architecture, the expected upper bound on its correlation with human ratings is approximated by
$ r_{\max} \approx \sqrt{r_{\text{human}}\cdot r_{\text{model}}}$, 
which is reported as $\mathrm{PLCC}$ ceiling in Table~\ref{tab:baseline_and_ceiling} (ranging here from ${\sim}0.74$ to ${\sim}0.78$). Practically, this implies that future improvements in predictive alignment will require (a) datasets for which human raters show stronger consensus on authenticity rating and/or (b) models whose predictions are more stable across training instantiations.

\begin{table}[t]
\centering
\setlength{\tabcolsep}{4pt}
\begin{tabularx}{\linewidth}{l *{5}{>{\centering\arraybackslash}X}}
\toprule
& \multicolumn{3}{c}{\textbf{Prediction metrics}} & \multicolumn{2}{c}{\textbf{Reliability}} \\
\cmidrule(lr){2-4}\cmidrule(lr){5-6}
\textbf{Architecture} &
\textbf{RMSE} &
\textbf{PLCC} &
\textbf{SRCC} &
\textbf{Model reliability} &
\textbf{PLCC Ceiling} \\
\midrule
Barlow Twins     & $\bm{7.10 \pm 0.11}$ & $0.62 \pm 0.01$ & $0.63 \pm 0.01$ & $0.91$ & $0.78$ \\
ResNet152       & $9.85 \pm 0.39$ & $0.51 \pm 0.02$ & $0.53 \pm 0.02$ & $0.91$ & $0.78$ \\
DenseNet161     & $7.48 \pm 0.22$ & $0.63 \pm 0.02$ & $0.62 \pm 0.02$ & $0.90$ & $0.78$ \\
EfficientNetB3  & $8.82 \pm 0.26$ & $0.50 \pm 0.02$ & $0.51 \pm 0.03$ & $0.84$ & $0.75$ \\
VGG16           & $8.62 \pm 0.43$ & $0.54 \pm 0.03$ & $0.57 \pm 0.03$ & $0.81$ & $0.74$ \\
VGG19           & $8.52 \pm 0.36$ & $0.56 \pm 0.03$ & $0.59 \pm 0.03$ & $0.81$ & $0.74$ \\
\bottomrule
\end{tabularx}
\caption{\textbf{Predictive performance and reliability across architectures.} Mean $\pm$ SD across 10 random seeds. Models were trained to minimize \textrm{MSE}, and best performance is bold-faced. \textrm{PLCC}: Pearson linear correlation coefficient. \textrm{SRCC}: Spearman rank correlation coefficient. Model reliability: mean pairwise Pearson correlation between prediction vectors across random initializations (45 pairs). \textrm{PLCC} ceiling: estimated upper bound on the observable correlation between a single trained instance and the human \textrm{MOS} vector, accounting for reliability of both human ratings and model predictions.}
\label{tab:baseline_and_ceiling}
\end{table}

\subsubsection{Separating authenticity-specific from quality-shared variance}

As indicated in the introduction, a general interpretive difficulty is that rated authenticity is strongly correlated with rated quality. This means that the strong performance achieved by some models may reflect learning covarying features related to quality, which would be sufficient to produce apparent ``authenticity prediction.'' To evaluate this, we first evaluated if $\hat{A}$ (authenticity prediction produced by the model) predicts quality ($Q$). Table~\ref{tab:partialCorrelations} reports the results. As expected, this was the case across architectures, with the correlation $r(Q,\hat{A})$ ranging from 0.47 to 0.65. Thus, a substantial component of the model's authenticity prediction tracks latent dimensions underlying quality judgments. Given this, we also evaluated if $\hat{A}$ will uniquely predict \textsc{authenticity} once the contribution of \textsc{quality} is removed. 

This analysis revealed statistically significant but relatively weak correlations that vary across architectures. Specifically, after partialling out $Q$ from both $A$ and $\hat{A}$, the residual correlation, $r(A,\hat{A}\mid Q)$, was in the range of 0.15--0.24 for Barlow Twins / ResNet152 / DenseNet161 / EfficientNetB3, and was near zero for the VGG architectures (see Table~\ref{tab:partialCorrelations}). Using the 10 variants per architecture as observations ($n=10$), a $t$-test indicated that the residual correlation was significant for Barlow Twins ($t(9)=25.30,\, p < 0.001$), ResNet152 ($t(9)=23.72,\, p < 0.001$), DenseNet161 ($t(9)=16.87,\, p < 0.001$), and EfficientNetB3 ($t(9)=21.08,\, p < 0.001$). However, it was not significant for VGG19 ($t(9)=1.36,\, p=0.21$) or VGG16 ($t(9)=0.53,\, p=0.61$). We conclude that, in the current dataset, most of the association between $\hat{A}$ and $A$ is explained by their shared variance with $Q$. However, a modest residual covariance remains for some architectures. This suggests that the models learn information beyond what drives quality judgments, but this authenticity-specific information is more limited.

The analysis also clarifies what underlies the relative performance of different models. For example, while Barlow Twins and DenseNet161 showed similar $\mathrm{PLCC}$ values, the former better predicted authenticity-specific information. Similarly, while DenseNet161 showed higher $\mathrm{PLCC}$ than EfficientNetB3 (0.63 vs. 0.50), as well as lower $\mathrm{RMSE}$ (see Table~\ref{tab:baseline_and_ceiling}), it was actually less effective at predicting the authenticity-specific component (0.16 vs. 0.20). Finally, the two VGG architectures appeared to rely primarily on quality-related variance. While the VGG-family models did not track authenticity-specific variance, we retain them in the subsequent analyses of explanations because they provide a potentially informative contrast class. If their explanations are consistent with those of architectures that do track authenticity, this would suggest that consistent explanations can be found between two architecture that track different types of information.

\begin{table}[t]
\centering
\setlength{\tabcolsep}{4pt}
\begin{tabularx}{\linewidth}{l *{3}{>{\centering\arraybackslash}X}}
\toprule
\textbf{Architecture} &
\textbf{$r(A,\hat{A})$} (PLCC) &
\textbf{$r(Q,\hat{A})$} &
\textbf{Partial $r(A,\hat{A}\mid Q)$} \\
\midrule

Barlow Twins   & $0.62 \pm 0.01$ & $0.60 \pm 0.02$ & $0.24^* \pm 0.03$ \\
ResNet152      & $0.51 \pm 0.02$ & $0.51 \pm 0.01$ & $0.15^* \pm 0.02$ \\
DenseNet161    & $0.63 \pm 0.02$ & $0.65 \pm 0.02$ & $0.16^* \pm 0.03$ \\
EfficientNetB3 & $0.50 \pm 0.02$ & $0.47 \pm 0.02$ & $0.20^* \pm 0.03$ \\
VGG19          & $0.56 \pm 0.03$ & $0.63 \pm 0.02$ & $0.03 \pm 0.07$ \\
VGG16          & $0.54 \pm 0.03$ & $0.61 \pm 0.03$ & $0.01 \pm 0.06$ \\

\bottomrule
\end{tabularx}

\caption{\textbf{The relation between authenticity predictions and quality ratings}. Correlations between true \textsc{Authenticity} scores ($A$), predicted \textsc{Authenticity} scores ($\hat{A}$), and  true \textsc{Quality} scores ($Q$) across model variants. Entries are mean $\pm$ SD across 10 variants per architecture. The partial correlation indicates the residual correlation between true and predicted \textsc{Authenticity} after partialling out $Q$ from both. Stars (*) indicate that the mean partial correlation differs significantly from zero (one-sample $t$-test against zero, $p < .05$).
}

\label{tab:partialCorrelations}
\end{table}

\subsection{Consistency of explanation maps within and across architectures}
\paragraph{Within-architecture consistency: Grad-CAM.}

The consistency of Grad-CAM-produced attributions (heatmaps) within architectures was high when trained from 10 random seeds (See Table \ref{tab:consistency_summary}). Consistency was also quite high when computed using Intersection-over-Union, with EfficientNetB3 and DenseNet161, showing slightly lower consistency than the other architectures. 

We also computed two additional statistics of within-model diversity.  The first was prediction-similarity, which we computed as the average pair-wise similarity of the predictions produced by each model's 10 variants. The second was representational similarity, which evaluated the consistency of representational geometry learned by the regression adapter. Here we considered an image's representation as its embedding on the final fully connected layer of the regression-head. From these we computed an object-by-object Representational Similarity Matrix \citep[RSM,][]{nili2014RSAtoolbox} for each of the 10 model variants, allowing us to compute representational-consistency within architecture.  The analyses demonstrate that Grad-CAM attribution maps provide information on model diversity not captured by these metrics. This can be best seen in Table~\ref{tab:consistency_summary}, examining the IoU@5\%, where DenseNet161 and EfficientNetB3 are indicated as providing markedly less consistent explanations. However, these two architectures produced prediction-similarity and RSM similarity values quite comparable to the other architectures. 

\begin{table}[t]
\centering
\setlength{\tabcolsep}{4pt}
\begin{tabularx}{\linewidth}{l *{6}{>{\centering\arraybackslash}X}}
\toprule
\textbf{Architecture} &
\textbf{Consistency (Corr.)} &
\textbf{IoU @ 5\%} &
\textbf{IoU @ 15\%} &
\textbf{IoU @ 25\%} &
\textbf{Pred. similarity (Corr.)} &
\textbf{RSM similarity (Corr.)} \\
\midrule

Barlow Twins   & $0.99 \pm 0.01$ & $0.77 \pm 0.13$ & $0.84 \pm 0.09$ & $0.88 \pm 0.07$ & $0.93$ & $0.78$ \\
ResNet152      & $0.99 \pm 0.01$ & $0.77 \pm 0.14$ & $0.85 \pm 0.10$ & $0.88 \pm 0.08$ & $0.91$ & $0.83$ \\
DenseNet161    & $0.78 \pm 0.20$ & $0.41 \pm 0.26$ & $0.52 \pm 0.21$ & $0.58 \pm 0.17$ & $0.90$ & $0.77$ \\
EfficientNetB3 & $0.85 \pm 0.14$ & $0.46 \pm 0.24$ & $0.56 \pm 0.19$ & $0.61 \pm 0.16$ & $0.85$ & $0.82$ \\
VGG19          & $0.99 \pm 0.01$ & $0.77 \pm 0.11$ & $0.83 \pm 0.07$ & $0.86 \pm 0.06$ & $0.84$ & $0.74$ \\
VGG16          & $0.99 \pm 0.01$ & $0.81 \pm 0.10$ & $0.86 \pm 0.06$ & $0.89 \pm 0.05$ & $0.84$ & $0.64$ \\

\bottomrule
\end{tabularx}
\caption{\textbf{Within-architecture explanation-consistency (Grad-CAM).}
Test-set Grad-CAM within-architecture consistency (mean $\pm$ SD) measured by correlations between attributions maps and using IoU at three saliency thresholds (top 5\%, 15\%, 25\% pixels). Prediction similarity: mean pairwise Pearson correlation between prediction vectors. RSM similarity: mean pairwise Pearson correlation between upper-triangle entries of cosine-similarity representational similarity matrices, computed from penultimate regression-head embeddings (test set).
}
\label{tab:consistency_summary}
\end{table}

\paragraph{Within-architecture consistency: LIME.}
LIME produces attribution maps by learning to approximate the effect of image perturbations on the original DNN model (Section~\ref{sec:lime}). In the current study, LIME's approximation of the original architectures (surrogate fidelity $R^2$) was moderate and  varied across architectures. It was highest on average for the two VGG architectures (Table \ref{tab:supmat_lime_r2_by_arch};  $R^2\approx 0.56, 0.60$) but lower for others (EfficientNetB3, ResNet152 $R^2 {\sim}0.43$). The moderate approximations mean that LIME may provide convergent evidence to those identified by MPM and Grad-CAM, but should be interpreted with caution.  

Nonetheless, the analysis of the LIME explanation maps provided convergent results to those identified by Grad-CAM. Within each architecture, explanation maps were similar across the 10 variants when evaluated via image-level correlations (mean pixel-level correlation $r {\sim}0.85$; see Table \ref{tab:lime_within}). The intersection over union (IoU) analysis also showed the same patterns found for Grad-CAM, with slightly lower consistency for DenseNet161 and EfficientNetB3 compared to the other architectures.  At the same time, IoU values were overall lower for LIME than for Grad-CAM. This could be due to the greater smoothness of Grad-CAM maps, where sampling from low-resolution feature maps can spread similar attribution values over large image areas, or the inherently greater noise in LIME which operates by using a surrogate model to measure the impact of perturbations.

We do not report within-architecture consistency for MPM as the computational cost of masking a single image across 60 variants renders this analysis highly intensive (on the order of several hours per image). Given the convergent within-architecture results from Grad-CAM and LIME, we treat those two methods as sufficient evidence for within-architecture consistency and use MPM for across-architecture and ensemble analyses.

\begin{table}[t]
\centering
\setlength{\tabcolsep}{4pt}
\begin{tabularx}{\linewidth}{l *{4}{>{\centering\arraybackslash}X}}
\toprule
\textbf{Architecture} &
\textbf{Consistency (Corr.)} &
\textbf{IoU @ 5} &
\textbf{IoU @ 15} &
\textbf{IoU @ 25} \\
\midrule
Barlow Twins   & $0.88 \pm 0.03$ & $0.52 \pm 0.13$ & $0.59 \pm 0.08$ & $0.63 \pm 0.07$ \\
ResNet152      & $0.86 \pm 0.04$ & $0.52 \pm 0.14$ & $0.57 \pm 0.08$ & $0.61 \pm 0.07$ \\
DenseNet161    & $0.85 \pm 0.03$ & $0.47 \pm 0.12$ & $0.53 \pm 0.07$ & $0.59 \pm 0.06$ \\
EfficientNetB3 & $0.76 \pm 0.06$ & $0.39 \pm 0.13$ & $0.45 \pm 0.08$ & $0.50 \pm 0.07$ \\
VGG16          & $0.85 \pm 0.03$ & $0.43 \pm 0.11$ & $0.51 \pm 0.07$ & $0.57 \pm 0.06$ \\
VGG19          & $0.83 \pm 0.04$ & $0.44 \pm 0.12$ & $0.51 \pm 0.07$ & $0.56 \pm 0.06$ \\
\bottomrule
\end{tabularx}
\caption{\textbf{Within-architecture explanation consistency (LIME).}
Mean $\pm$ SD across images of within-architecture Pearson's correlation (mean pairwise across variants) and IoU overlap between the top-$K$ most positive LIME coefficients.}
\label{tab:lime_within}
\end{table}

\begin{figure}[t]
    \centering
    \includegraphics[width=1\linewidth]{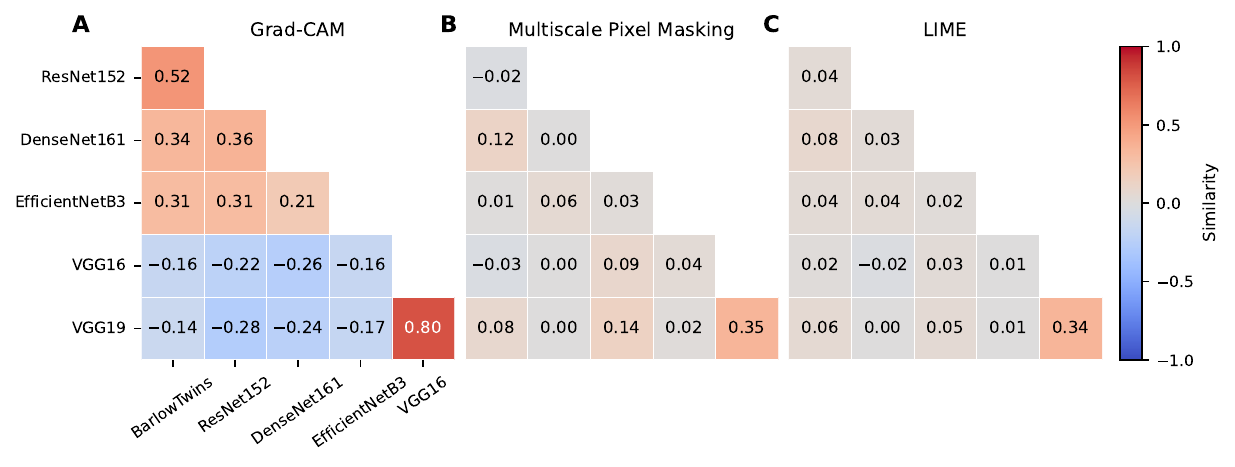}
    \caption{\textbf{Across-architecture explanation agreement.} Each cell shows the mean Spearman correlation between prototype explanation maps for a pair of architectures, averaged across test-set images. Panel A: Grad-CAM. Panel B: Multiscale Pixel Masking (MPM). Panel C: LIME. Higher values indicate greater similarity. SDs across images are provided in Appendix Figure~ \ref{fig:supmatAcrossArchConsistency}.}
    \label{fig:consacross}
\end{figure}

\paragraph{Across-architecture agreement: Grad-CAM.} 
As shown in Figure \ref{fig:consacross}A, correlations between image-level explanations tended to be moderate between architectures, and much lower than the within-architecture consistency values. In particular, DenseNet161, EfficientNetB3, ResNet152 and Barlow Twins tended to show positive correlations. The two VGG architectures produced different explanations, consistent with our findings indicating they track different information.  These patterns maintained when consistency was evaluated using Intersection-over-Union (see Appendix Figure \ref{fig:supmatAcrossArchConsIOU}). 

\begin{figure}[t]
        \centering
        \includegraphics[width=1\linewidth]{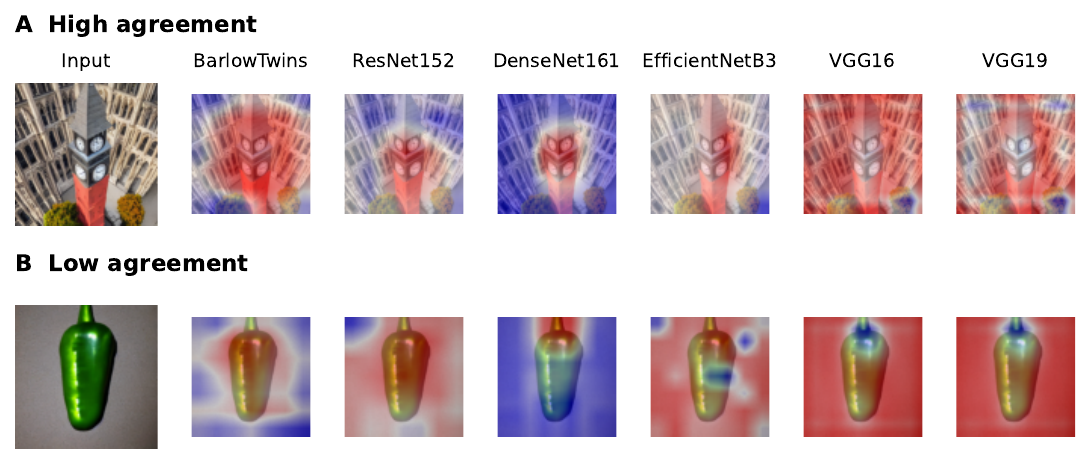}
        \caption{\textbf{Examples of across-architecture Grad-CAM agreement.} Each column shows the attribution map produced by one architecture for the same image. Upper row: an image with relatively high across-architecture agreement. Lower row: an image with low across-architecture agreement. Note that VGG architectures systematically produced attribution maps that differed from the other architectures (see text).}
        \label{fig:betweenarch}
\end{figure}

While the average correlation was, on average, low to moderate, the range across images was substantial, with some images showing strong attribution-correlations across architectures, and others weak correlations (for range, see Appendix Figure \ref{fig:supmatAcrossArchConsistency}). An example of images with relatively high or low across-architecture consistency is given in Figure \ref{fig:betweenarch}.

\paragraph{Across-architecture agreement: MPM and LIME.} For MPM, correlations of attribution maps across architectures were, on average, near zero (Figure \ref{fig:consacross}B). The only exception were the two VGG architectures where a moderate correlation was found (Pearson's $r = 0.35$). The low correlations do not necessarily mean that the procedure itself lacks explanatory validity. In fact, there were some images for which MPM attribution maps were quite similar across models, with $r>0.5$. However, it could be that the pixel-level value assignment and much lower smoothing factor associated with MPM limited the strength of correlation.  The LIME analysis converged with the MPM findings, showing relatively low correlations overall, with the strongest for the VGG architectures.

\paragraph{Relation between within-architecture attribution consistency and authenticity.}~We first computed this analysis for the test-set images (Table \ref{tab:consistency_human}). Recall that for purposes of creating attribution maps, pruning optimized the adapter's readout for the test set. For two architectures, Barlow Twins and EfficientNetB3, we found statistically significant positive correlations between explanation consistency and human authenticity ratings. For EfficientNetB3 this positive correlation was found independent of how consistency was computed (correlation- or IoU-based computations). For Barlow Twins, it was found when consistency was computed using a correlation-based approach across explanations. Interestingly, when computing these correlations for the train-set data, positive correlations were even more prevalent (even though the readout was optimized for the test-set), and found for Barlow Twins, ResNet152, DenseNet161 and EfficientNetB3, across multiple methods of computing consistency. These findings show that the relative consistency of attributions is not random, but is related to image content that drives human authenticity judgments.

\begin{table}[t]
\centering
\setlength{\tabcolsep}{2pt}
\begin{tabularx}{\linewidth}{l *{8}{>{\centering\arraybackslash}X}}
\toprule
\textbf{Architecture} &
\multicolumn{4}{c}{\textbf{Test}} &
\multicolumn{4}{c}{\textbf{Train}} \\
\cmidrule(lr){2-5}\cmidrule(lr){6-9}
 & Corr.-Based & Top 5\% & Top 15\% & Top 25\%
 & Corr.-Based & Top 5\% & Top 15\% & Top 25\% \\
\midrule

Barlow Twins
& \textbf{0.31*} & 0.11 & 0.15 & 0.13
& \textbf{0.25*} & \textbf{0.16} & \textbf{0.22} & \textbf{0.26*} \\

ResNet152
& 0.12 & 0.11 & 0.11 & 0.12
& \textbf{0.08} & \textbf{0.11} & \textbf{0.16} & \textbf{0.16} \\

DenseNet161
& 0.01 & $-0.12$ & $-0.03$ & 0.07
& \textbf{0.11} & \textbf{0.10} & \textbf{0.12} & \textbf{0.14} \\

EfficientNetB3
& \textbf{0.25*} & \textbf{0.21} & \textbf{0.25*} & \textbf{0.29*}
& \textbf{0.24*} & \textbf{0.14} & \textbf{0.20} & \textbf{0.23*} \\

VGG19
& 0.05 & $-0.01$ & 0.08 & 0.09
& 0.03 & $-0.02$ & 0.01 & 0.05 \\

VGG16
& 0.04 & 0.06 & 0.10 & 0.06
& 0.01 & 0.02 & 0.05 & \textbf{0.07} \\

\bottomrule
\end{tabularx}

\caption{
\textbf{Images perceived as more authentic are associated with more consistent explanations}. Pearson correlations between explanation consistency and mean human authenticity ratings. Train and test sets are reported separately. Bold values indicate $p < .05$ (uncorrected). Stars (*) indicate correlations that survive FDR correction for 48 multiple comparisons ($p < .0067$).
}
\label{tab:consistency_human}
\end{table}

\paragraph{Relation between explanation consistency and image-level prediction accuracy.}
We found only weak, non-significant correlations between explanation consistency and prediction error across all IoU metrics (Appendix Table~\ref{tab:consistency_vs_error}). While EfficientNetB3 ($r=-0.141$ at top-5\% IoU) and VGG19 ($r=-0.141$ at top-15\% IoU) showed the expected inverse relationships, these remained low. 

\subsection{Ensemble prediction and attribution}
As shown in Table \ref{tab:ensemble_accuracy}, ensemble methods performed better than any single model. Bagging, which averaged the predictions (scalar regression outputs) across all 60 independent pruned variants, showed the best performance, both in terms of $\mathrm{RMSE}$ and correlations with human ratings. Stacking also outperformed all single models.  A direct comparison between Bagging and Stacking is not reported, as the two methods were evaluated on different held-out sets (see Table \ref{tab:partition_roles}).

\begin{figure}[t]
    \centering
    \includegraphics[width=1\linewidth]{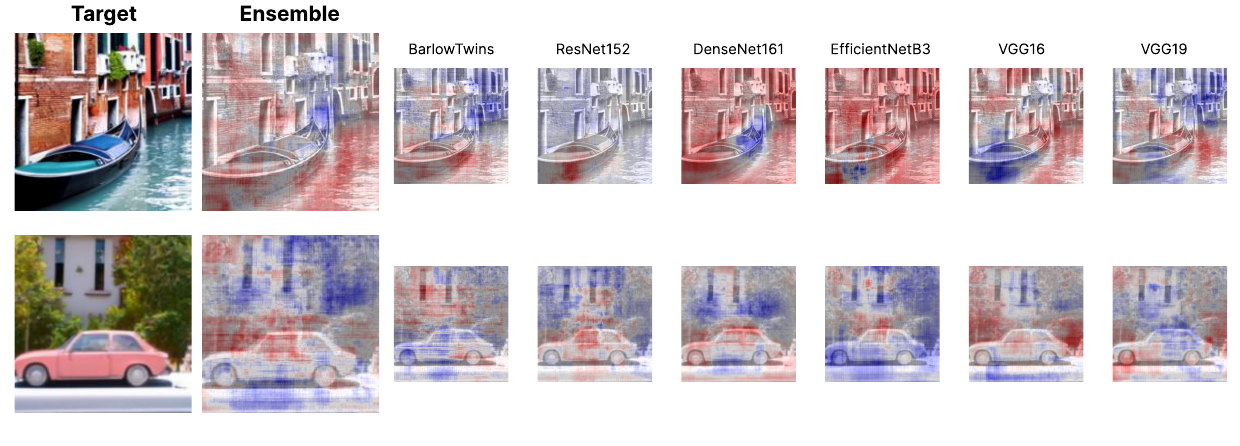}
    \caption{\textbf{MPM attributions provided for Ensemble explanations.} For two target images, MPM attribution maps are shown for the Bagging Ensemble and for each of the six individual architectures. Single-architecture maps are averaged across 10 model variants. Warm colors indicate regions whose occlusion reduces predicted authenticity; cool colors indicate regions whose occlusion increases it.}
    \label{fig:ensemble_mpm}
\end{figure}

Using MPM we could  produce attribution maps for the Ensemble models themselves. Figure \ref{fig:ensemble_mpm} shows examples of such attributions alongside those produced by the single models. The better predictive performance appears to be accompanied by attribution maps that spatially segregate image elements associated with higher or lower perceived authenticity. 

We note that a limitation of this method is that producing an explanation for a single image in the current setup requires several hours as the impact of each mask is evaluated across all 10 variants in each of the six architectures. To address this issue, we evaluated if LIME surrogate models could approximate the ensemble's encoding function. (i.e., whether the LIME model's predictions for perturbed versions of the original image would track the Ensemble model's predictions for those same perturbations). We found that although the LIME surrogates achieved reasonable predictive alignment with the ensemble ($R^2$ fits $\approx 0.6-0.7$), the attribution maps produced by LIME showed only moderate similarity with ensemble-derived maps ($r \approx 0.5–0.6$). We conclude that for the current dataset and model family, LIME is not a viable surrogate for Ensemble attribution.

\begin{table}[t]
\centering
\setlength{\tabcolsep}{5pt}
\begin{tabularx}{\linewidth}{l >{\centering\arraybackslash}X *{2}{>{\centering\arraybackslash}X}}
\toprule
\textbf{Architecture} &
\textbf{RMSE (mean $\pm$ SD) [best variant]} & \textbf{PLCC} & \textbf{SRCC} \\
\midrule
Barlow Twins    & $7.01 \pm 0.25$ [6.65] & $0.64 \pm 0.03$ & $0.65 \pm 0.03$ \\
ResNet152       & $9.14 \pm 0.31$ [8.71] & $0.48 \pm 0.03$ & $0.50 \pm 0.03$ \\
DenseNet161     & $7.43 \pm 0.25$ [6.95] & $0.63 \pm 0.02$ & $0.62 \pm 0.02$ \\
EfficientNetB3  & $8.89 \pm 0.33$ [8.33] & $0.47 \pm 0.01$ & $0.48 \pm 0.01$ \\
VGG16           & $8.57 \pm 0.16$ [8.29] & $0.52 \pm 0.02$ & $0.55 \pm 0.02$ \\
VGG19           & $8.37 \pm 0.35$ [7.79] & $0.55 \pm 0.03$ & $0.59 \pm 0.03$ \\
\midrule
\textbf{Bagging Ensemble}  & \textbf{6.04} & \textbf{0.73} & \textbf{0.73} \\
\textbf{Stacking Ensemble} & 6.14          & 0.71          & 0.70          \\
\bottomrule
\end{tabularx}
\caption{
\textbf{Ensembles outperform all individual models.} For base architectures, $\mathrm{RMSE}$ is reported as mean $\pm$ SD across 10 unpruned variants, with the best pruned variant (lowese \textrm{RMSE}) in brackets. $\mathrm{PLCC}$ and $\mathrm{SRCC}$ are mean $\pm$ SD across the ten variants. All results on held-out data.
}
\label{tab:ensemble_accuracy}
\end{table}

\section{Discussion}

Deep neural networks have recently been shown to predict human perceptions of authenticity with good accuracy \citep{wang2023aigciqa2023, chen2024Natural}. Understanding the mechanisms that underlie this performance is increasingly important given the proliferation of deepfake images, which in some domains cannot be reliably differentiated from real images \citep[e.g.,][]{nightingale2022ai}. We therefore evaluated whether vision DNNs can be used to produce meaningful explanations of their predictions of human authenticity judgments. 

\paragraph{Pretrained models vary in predictive capacity and the information tracked.}
Authenticity ratings in the studied dataset had good split-half reliability, with an estimated noise ceiling of $\approx0.82$, and were therefore a valid target for prediction. Some of the pre-trained architectures achieved moderately strong alignment with human authenticity ratings, with the best-performing models reaching mean $\mathrm{PLCC} \approx 0.63$ (about 80\% of the noise ceiling). Other architectures performed substantially worse. 

An initial finding was that different models track different decision-relevant information. In particular, the two VGG-based models (VGG16, VGG19) learned to predict the authenticity target variable by tracking quality-related, rather than authenticity-related, image cues. When quality information was partialled out, these models did not account for any remaining variance in the authenticity ratings they were trained to predict. Indeed, their authenticity predictions on held-out data were more strongly correlated with quality ratings than with authenticity ratings. In contrast, EfficientNetB3, while performing similarly to these models, did account for variance unique to authenticity. Thus, not all models that predict authenticity are tracking authenticity-specific cues. Our findings also indicate an interesting difference between predicting human authenticity judgments for images and predicting human brain activity elicited by images. For the latter, different computer-vision architectures have been shown to produce similarly accurate predictions of representational responses in human inferior temporal cortex, leading the authors to suggest that ``the commonalities among these diverse DNNs appear to matter more than their differences, when considered as models of human vision'' \citep[][]{storrs2021diverse}. In contrast, we find substantial difference in predictive performance with the best model (Barlow Twins) producing $\mathrm{RMSE} = 7.10$ and worst model (ResNet152) $\mathrm{RMSE} = 9.85$. Considering that $\mathrm{RMSE}$ is matched to units of the original scale, Barlow Twins is on average more accurate by 2.7 scale points.  

\paragraph{Explanation consistency is a necessary (but not sufficient) condition for psychological relevance.}
Our main aims involved evaluating the consistency of explanation maps, which we take to be an essential criterion for using DNN interpretability tools in psychological contexts. Explainability methods are often treated as providing direct insight into what information a model uses, but their construct validity is itself a topic of ongoing discussion. We use Grad-CAM, which has been shown to pass important sanity checks \citep{adebayo2018sanity}, and a pixel-masking approach \citep{zeiler2014visualizing, palazzo2020decoding} which is one of the only black-box approaches for producing explanations from computer-vision models. Using frozen, pre-trained architectures, we trained a regression head as an adapter for readout. We evaluated if the resulting models provided similar explanations for the same image when the adapter was retrained from different random initializations. As an initial sanity check we evaluated whether these architectures produced consistent scalar predictions across training runs, as this is a requirement towards producing consistent explanations. We quantified reliability by computing the similarity of prediction vectors across models trained from 10 random seeds and found values ranging from 0.81 to 0.91, indicating good reliability.

Analyzing the attributions produced, we find that readout from some architectures (including Barlow Twins, VGG variants, and ResNet) is associated with high within-architecture consistency, when consistency is defined as pixel-wise correlations between attribution heatmaps. A more stringent criterion that quantifies agreement among salient regions (intersection over union) indicated relatively high consistency for those architectures, but lower consistency for DenseNet and EfficientNet.  Furthermore, explanation consistency covaried with perceived authenticity such that images rated as more authentic were associated with more consistent attribution maps across seeds, particularly for EfficientNetB3 and Barlow Twins (Table \ref{tab:consistency_human}). Importantly, this relationship was non-existent for the VGG architectures, where authenticity prediction was dominated by quality-related variance.

Taken together, these findings suggest that highly authentic images contain evidence that for some architectures ends up being consistently encoded (across seeds) along feature-space directions read out by the regression head. In contrast, low-authenticity images may be compatible with different features that are coded by different readout directions across runs.  Our findings are consistent with prior work showing that the same architecture trained from different seeds can produce different explanations \citep{watson2022agree}. 

Notwithstanding, the main obstacle to psychological interpretation in our data is the low across-architecture consistency of explanations. Architectures with good levels of within-architecture consistency still produced different attribution maps for the same images. This corresponds to previously described Rashomon-type results which refer to cases where similarly performing models provide different explanations \citep[e.g.,][]{muller2023empirical, gwinner2024comparing}. Here we focus on the psychological implication, which is that if multiple models predict human ratings well but highlight different image evidence, it is then difficult to justify treating any single model’s explanations as reflecting the cues that drive human judgments.  These findings have direct relevance to how explanations can be used in applied settings. For example, \citet{boyd2023value} found that when people discriminate real from fake images, presenting them with CAM-based explanations from an AI system does not shift their decisions. 

Of course, our results do not suggest that all explanations are wrong. For example, one of the architectures might provide more valid explanations, or perhaps different architectures provide complementary information on authenticity cues driving human judgments. In both cases, the practical utility of explanations provided by any single model is reduced unless multiple models are evaluated in parallel.  As a result, establishing which explanation (if any) is more  psychologically valid would require cross-referencing against external human benchmarks. That said, behaviorally identifying image elements that drive authenticity perception is far from trivial: tasks that ask observers to indicate ``authentic'' regions presuppose that the relevant cues are spatially localizable, as do behavioral paradigms based on local masking, and gaze-tracking methods may be driven by saliency or other image qualities.

From a methodological perspective, one can ask whether it is necessary to quantify explanation consistency at all, given that consistency in model performance (prediction consistency) or in representations (RSM consistency) might provide similar information. Here we find important dissociations. For example, VGG variants exhibited relatively high explanation consistency but showed lower similarity in representational geometry (RSMs computed from penultimate-layer embeddings of the regression model). This is expected because RSMs capture the \textit{geometry} of stimulus representations, but attribution maps emphasize \textit{task-relevant directions} in representation space. Variation in RSM structure can reflect representational dimensions that are not important for the regression output. The relation to prediction-similarity was less clear and unexpected: For example, EfficientNetB3 produced highly similar predictions across runs for the hold-out set, but relatively dissimilar explanations. 

\paragraph{Ensembles improve prediction and enable attribution.}
The across-architecture diversity of explanations motivated us to produce a solution in which information is combined across multiple architectures. In our data, ensemble predictors produced the highest accuracy and the strongest alignment with human judgments. In fact, ensembles often perform best when different models produce diverse predictions for the same dataset \citep[e.g.,][]{brown2005diversity}, and therefore provide a natural response to the Rashomon effect we document. Because different architectures capture non-identical evidence, aggregating information can produce more accurate predictions when computed for the ensemble output.

In related work, it has been shown that diversity in explanations can be used to construct a consensus explanation, and that models closer to such a consensus can exhibit improved performance \citep{li2023crossRashomon}. While we do not construct consensus explanations here, we compute explanations for a consensus \emph{predictor} (i.e., the ensemble model). Future work can evaluate the relation between these two forms of explanation.

\paragraph{DNNs as explanatory models: the role of explanation-consistency}
Since the initial applications of DNNs to modeling behavior, there has been substantial debate about their utility as models of human cognition. On one extreme, some have argued that successful prediction alone should be taken as a useful starting point for generating hypotheses about human behavior \citep[e.g.,][]{cichy2019deep}. On the other, some have argued that although DNNs can predict behavior and neural activity, they fail on many perceptual tasks, and thus there is little reason to think they instantiate representations similar to those in humans \citep[e.g.,][]{bowers2023deep}. We take an intermediate position: DNNs may be useful for understanding human cognition if it is possible to obtain explanations for their predictions and decisions that are relevant to the target behavior.

The relevance of DNNs is very likely task-dependent. It has been shown that for some tasks, a pre-trained network's feature space provides a good approximation for human behavior that further improves with fine-tuning \citep[e.g., similarity judgments; ][]{peterson2018evaluating, tarigopula2023improved}. For other tasks, such as object discrimination, the correspondence between DNN and human errors at the single-image level can be quite low \citep{rajalingham2018large}. In the current study, we find that model predictions are satisfactory, and their explanations often consistent within architecture. 

A recurring objection is that even when their behavior is similar to humans, DNNs may achieve it by tracking information that differs from the information humans use, but happens to covary with it. For example, ImageNet-trained CNNs have been argued to be biased toward texture \citep{geirhos2018imagenet} (but see \citet{burgert2025imagenet}). Our data show that this concern may be justified, but it is not fatal. Specifically, in support of said objection we find that VGG architectures indeed predict human authenticity ratings by tracking quality-related image dimensions: when quality is partialled out from the target variable, these models do not account for the remaining variance. However, other models did track authenticity-related information. This clearly shows that while DNN--human alignment can arise for different reasons, it is still possible to clarify what information a model is actually using when trained on authenticity ratings.

\paragraph{Limitations and future directions.}
Several limitations follow from the study. First, prediction performance is constrained by the noise ceiling of the human data. The human noise ceiling is an upper bound on achievable $\mathrm{PLCC}$ from a model, and as discussed, seed reliability further bounds this value. While the human noise ceiling was relatively high ($\approx 0.82$), better data-collection procedures could, in principle, raise it further. Because prediction of human ratings is only moderate (with correlations typically below 0.7), even stable explanations should be treated cautiously as candidates for human processing. 

Second, we did not fine-tune the backbones themselves, but instead attached a regression head. We made this choice because of the small size of the dataset (${\sim}950$ images per fold), as fine-tuning the backbone would require substantially more data than fitting an adapter. For the different architectures, the regression head consisted of 500K to ${\sim}1000K$ (Table \ref{tab:app_RegParamCount}) and adapting the backbone itself could increase the number of parameters by an order of magnitude in some cases.  The limitation is that because backbones are fixed, consistency does not reflect stability of the architecture itself across training. Instead, it reflects stability of the readout head. That is, it indicates if regression heads trained from random initializations, and on different train/validation splits select the same directions from the fixed representational space. High consistency means the head picks up on authenticity-related factors that are picked across different runs. Finally, regarding the internal validity of explanations, we did not perform formal sanity/fragility analyses for the explanation method within each architecture \citep{adebayo2018sanity}. 

\paragraph{Conclusions.} Our findings suggest that while multiple architectures can predict human authenticity judgments, they may do so for different reasons. Moreover, architectures can produce very  different explanations for the same images even when their explanations are internally consistent across random initializations. Together, these results indicate that, at present, predictive success alone does not provide sufficiently strong evidence for treating any single architecture as an explanatory model of human authenticity perception. At the same time, our results indicate a path for future work. They suggest a modeling framework in which prediction accuracy and explanation consistency are separable evaluation dimensions, and where ensembles are used as a possible solution when cross-architecture consensus in explanations is lacking.

\newpage

\section*{Statements and Declarations}
\paragraph{Funding.} The authors have no relevant financial interests to disclose.

\paragraph{Competing interests.} The authors have no competing interests to disclose.

\paragraph{Availability of data and material.}
The AIGCIQA2023 dataset used in the current study is available at 
\href{https://github.com/wangjiarui153/aigciqa2023}{GitHub}.

\paragraph{Code availability.}
Code for performing the core experiments is available at
\href{https://github.com/icaro-rdp/Internship_project_2025}{GitHub}.

\paragraph{Authors' contributions.} IrD: Conceptualization, Methodology, Software, Investigation, Data Analysis, Writing – original draft.
UH: Conceptualization, Methodology, Software review, Validation, Supervision, Data Analysis, Writing – review \& editing, Writing – final draft. The work was based in part on the first author’s master’s thesis.

\paragraph{Ethics approval.} Not applicable.
\paragraph{Consent to participate.} Not applicable.
\paragraph{Consent for publication.} Not applicable.

\bibliography{iclr2025_conference}
\bibliographystyle{iclr2025_conference}

\clearpage

\appendix
\counterwithin{figure}{section}
\counterwithin{table}{section}
\section{Appendix}
\subsection{Supplementary Figures}
\begin{figure}[H]
    \centering
    \includegraphics[width=0.5\linewidth]{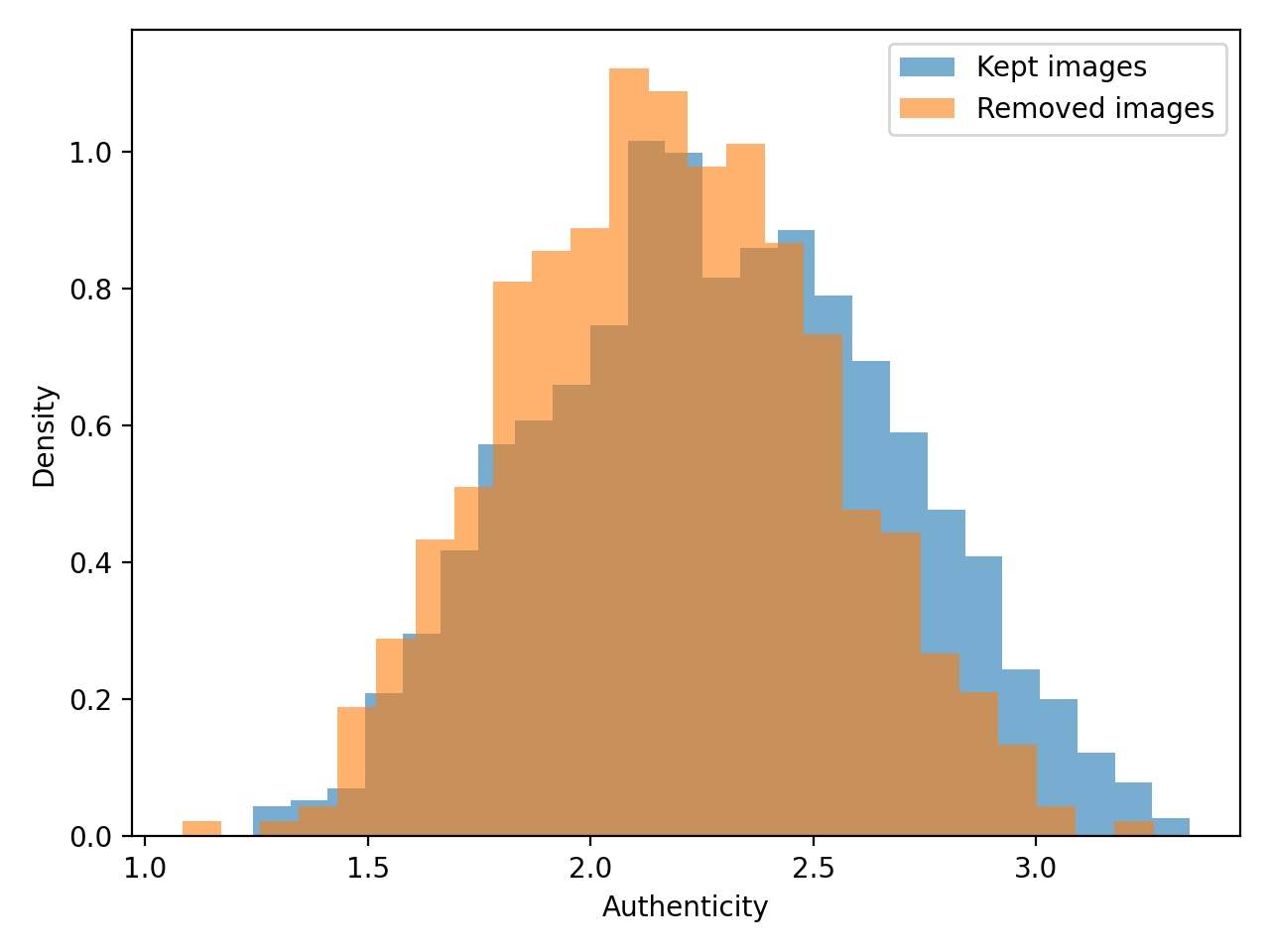}
    \caption{\textbf{Authenticity score distributions for excluded versus retained images.} Per-image mean \textsc{authenticity} scores for images excluded (\textit{Removed}, orange; $n=1033$) and retained (\textit{Kept}, blue; $n=1367$) based on prompt-metadata criteria. Removed images received significantly lower authenticity ratings (see text).}
    \label{fig:AppRemovedImages}
\end{figure}

\begin{figure}[H]
    \centering
    \includegraphics[width=1\linewidth]{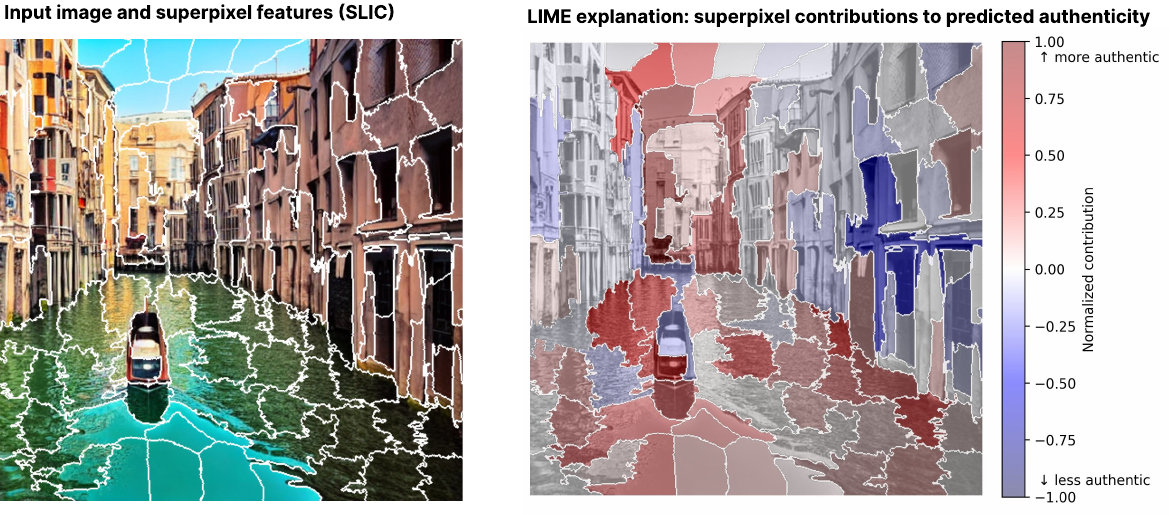}
    \caption{\textbf{LIME explanation of authenticity judgments} Left: input image with SLIC superpixel boundaries overlaid. In the LIME workflow, each superpixel is treated as a binary feature. In each perturbation, a random subset of superpixels is masked (replaced by the image mean RGB) and the resulting image is passed through the model for inference. Right: LIME explanation heatmap showing the contribution of each superpixel to the model's predicted authenticity score. Red: superpixel presence increases predicted authenticity. Blue: superpixel presence decreases predicted authenticity.}
    \label{fig:supmatLIME}
\end{figure}

\subsubsection{Across-architecture consistency with SD}
\begin{figure}[H]
    \centering
    \includegraphics[width=1\linewidth]{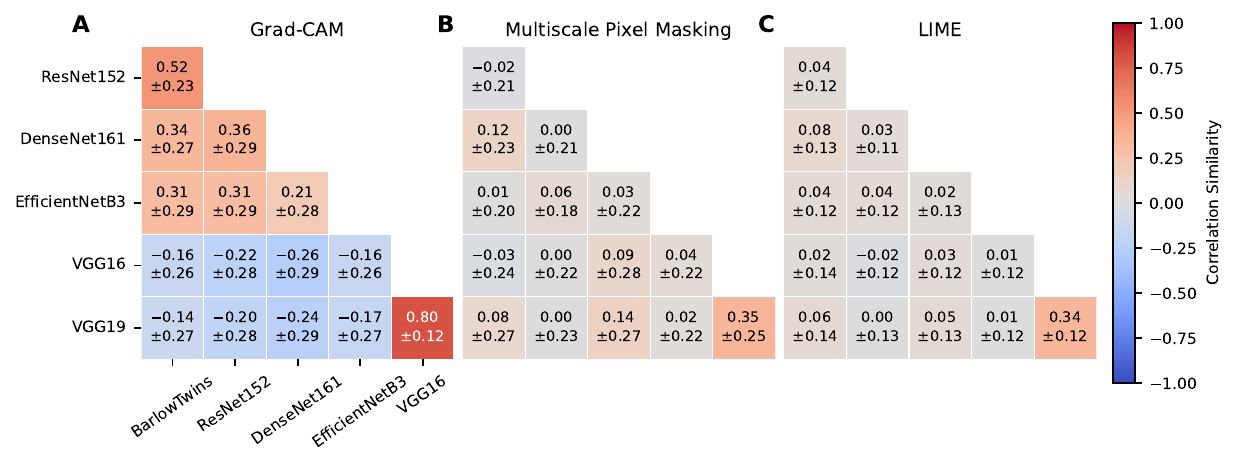}
    \caption{\textbf{Across-architecture Grad-CAM agreement with variability.} Supplementary to Figure~\ref{fig:consacross}. Each cell shows the mean Spearman correlation between prototype Grad-CAM maps for a pair of architectures, averaged across test-set images, with the SD across images in parentheses.}
    \label{fig:supmatAcrossArchConsistency}
\end{figure}

\begin{figure}[H]
    \centering
    \includegraphics[width=1\linewidth]{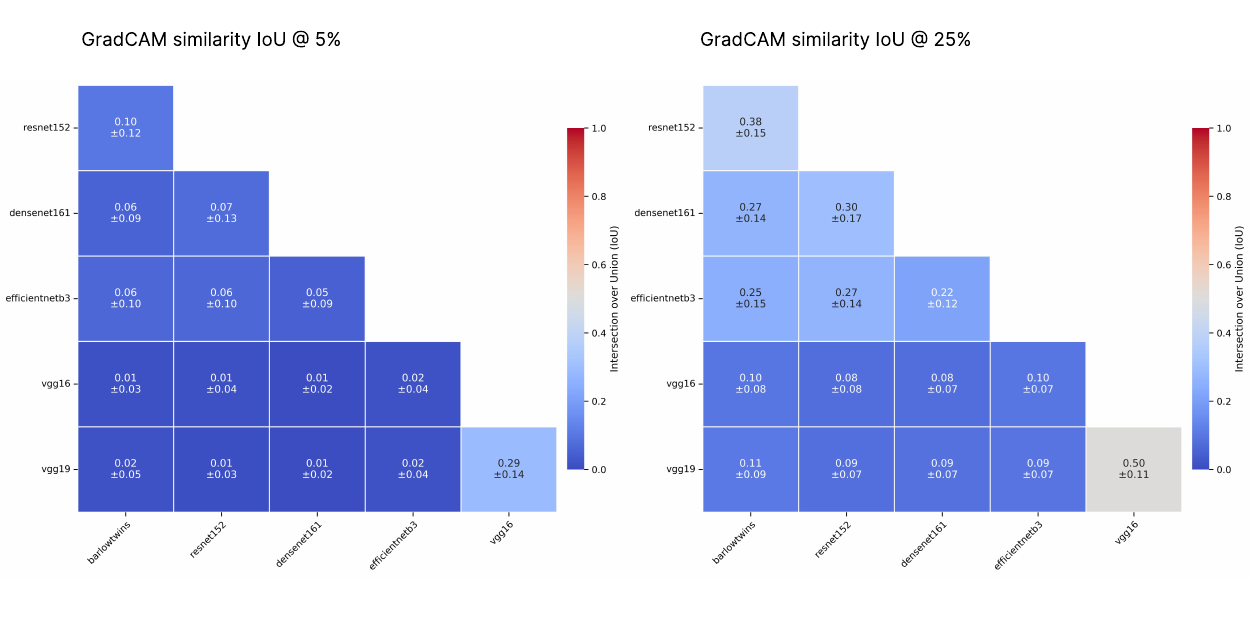}
    \caption{\textbf{Across-architecture similarity (IoU)}. Each cell shows average Intersection-Over-Union overlap of top 5\% (and top 25\%) pixels over images with standard deviation over images.}
    \label{fig:supmatAcrossArchConsIOU}
\end{figure}

\begin{figure} [H]
    \centering
    \includegraphics[width=1\linewidth]{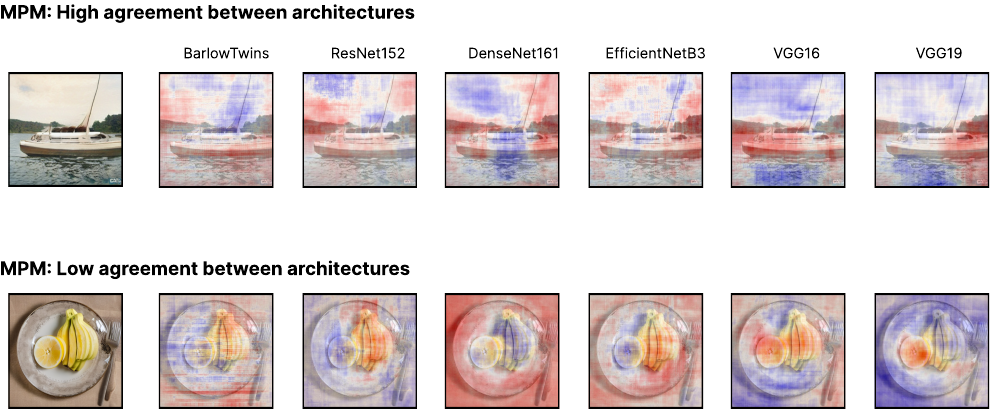}
    \caption{\textbf{Examples of across-architecture MPM agreement.} Each row shows MPM attribution maps for the same image across all six architectures. Upper rows: images with relatively high across-architecture agreement. Lower rows: images with low agreement.}
    \label{fig:supmatMPMAcrossArchCons}
\end{figure}

\newpage
\subsection{Supplementary Tables}

\begin{table}[h]
    \centering
    \begin{tabular}{ll}
        \toprule
        \textbf{Model} & \textbf{Target Layer} \\
        \midrule
        Barlow Twins & \texttt{features.7.2.conv3} \\
        ResNet152 & \texttt{features.7.2.conv3} \\
        DenseNet161 & \texttt{features.denseblock4.denselayer24.conv2} \\
        EfficientNetB3 & \texttt{features.8.0} \\
        VGG16 & \texttt{features.28} \\
        VGG19 & \texttt{features.34} \\
        \bottomrule
    \end{tabular}
    \caption{\textbf{Target convolutional layers for SBS pruning and Grad-CAM attribution.}}
    \label{tab:pruning_target_layers}
\end{table}

\begin{table}[h]
\centering
\begin{tabular}{l c c}
\hline
Architecture & $R^2$ (mean$\pm$SD) & min--max \\
\hline
Barlow Twins        & $0.537 \pm 0.091$ & 0.243--0.816 \\
DenseNet161        & $0.478 \pm 0.081$ & 0.274--0.773 \\
EfficientNetB3     & $0.437 \pm 0.076$ & 0.226--0.773 \\
ResNet152          & $0.439 \pm 0.083$ & 0.205--0.794 \\
VGG16              & $0.603 \pm 0.077$ & 0.289--0.820 \\
VGG19              & $0.562 \pm 0.079$ & 0.302--0.787 \\
\hline
\end{tabular}
\caption{\textbf{LIME surrogate fidelity by architecture.} For each architecture, we report the locality-weighted $R^2$ of the surrogate fit vs. the predictions of the original model, aggregated across all explained images and variants.}
\label{tab:supmat_lime_r2_by_arch}
\end{table}

\begin{table}[t]
\centering
\begin{tabular}{l l r}
\hline
\textbf{Backbone / model} & \textbf{Regression head} & \textbf{\# Reg head. params} \\
\hline
VGG16 & $4096 \rightarrow 128 \rightarrow 1$ & 524{,}545 \\
VGG19 & $4096 \rightarrow 128 \rightarrow 1$ & 524{,}545 \\
EfficientNet-B3 & $1536 \rightarrow 512 \rightarrow 128 \rightarrow 1$ & 852{,}737 \\
DenseNet-161 & $2208 \rightarrow 512 \rightarrow 128 \rightarrow 1$ & 1{,}196{,}801 \\
ResNet-152 & $2048 \rightarrow 512 \rightarrow 128 \rightarrow 1$ & 1{,}114{,}881 \\
Barlow Twins (ResNet50 encoder) & $2048 \rightarrow 512 \rightarrow 128 \rightarrow 1$ & 1{,}114{,}881 \\
\hline
\end{tabular}
\caption{\textbf{Regression head architectures and trainable parameter counts.} Each head maps the frozen backbone's output to a scalar authenticity score. }
\label{tab:app_RegParamCount}
\end{table}

\begin{table}[t]
\centering
\setlength{\tabcolsep}{4pt}
\begin{tabularx}{\linewidth}{l *{4}{>{\centering\arraybackslash}X}}
\toprule
\textbf{Architecture} &
\textbf{Consistency (Corr.)} &
\textbf{IoU @ 5\%} &
\textbf{IoU @ 15\%} &
\textbf{IoU @ 25\%} \\
\midrule

Barlow Twins   & $0.10$  & $-0.02$ & $-0.02$ & $0.02$ \\
ResNet152      & $-0.01$ & $-0.04$ & $-0.04$ & $-0.09$ \\
DenseNet161    & $-0.08$ & $-0.05$ & $-0.09$ & $-0.12$ \\
EfficientNetB3 & $-0.13$ & $-0.14$ & $-0.11$ & $-0.08$ \\
VGG19          & $0.04$  & $-0.06$ & $-0.12$ & $-0.03$ \\
VGG16          & $-0.01$ & $0.05$  & $0.08$  & $0.01$ \\

\bottomrule
\end{tabularx}

\caption{\textbf{Explanation consistency does not reliably predict image-level prediction error.} Pearson correlations between per-image Grad-CAM explanation consistency and mean absolute prediction error (MAE). IoU-based consistency at top-5\%, 15\%, and 25\% saliency thresholds. All correlations are non-significant.
}
\label{tab:consistency_vs_error}
\end{table}

\begin{table}[]
\centering
\scriptsize
\setlength{\tabcolsep}{4pt}  
\begin{tabularx}{\linewidth}{l *{3}{c} *{3}{c} c}
\toprule
\textbf{Architecture} &
\multicolumn{3}{c}{\textbf{Baseline}} &
\multicolumn{3}{c}{\textbf{Pruned}} &
\textbf{Red. (\%)} \\
\cmidrule(lr){2-4}\cmidrule(lr){5-7}
 & RMSE & PLCC & SRCC & RMSE & PLCC & SRCC &  \\
\midrule

Barlow Twins
& $7.10 \pm 0.11$ & $0.62 \pm 0.01$ & $0.63 \pm 0.01$
& $\bm{5.29 \pm 0.20}$ & $\bm{0.80 \pm 0.02}$ & $\bm{0.80 \pm 0.02}$
& $28.15 \pm 4.06$ \\

ResNet152
& $9.85 \pm 0.39$ & $0.51 \pm 0.02$ & $0.53 \pm 0.02$
& $8.53 \pm 0.34$ & $0.54 \pm 0.03$ & $0.55 \pm 0.03$
& $8.23 \pm 0.68$ \\

DenseNet161
& $7.48 \pm 0.22$ & $0.63 \pm 0.02$ & $0.62 \pm 0.02$
& $7.13 \pm 0.18$ & $0.65 \pm 0.02$ & $0.64 \pm 0.02$
& \textbf{$40.83 \pm 4.30$} \\

EfficientNetB3
& $8.82 \pm 0.26$ & $0.50 \pm 0.02$ & $0.51 \pm 0.03$
& $7.56 \pm 0.39$ & $0.61 \pm 0.03$ & $0.61 \pm 0.03$
& $8.20 \pm 0.78$ \\

VGG16
& $8.62 \pm 0.43$ & $0.54 \pm 0.03$ & $0.57 \pm 0.03$
& $7.43 \pm 0.30$ & $0.63 \pm 0.02$ & $0.65 \pm 0.02$
& $28.14 \pm 3.42$ \\

VGG19
& $8.52 \pm 0.36$ & $0.56 \pm 0.03$ & $0.59 \pm 0.03$
& $7.23 \pm 0.27$ & $0.65 \pm 0.02$ & $0.67 \pm 0.02$
& $25.21 \pm 4.64$ \\

\bottomrule
\end{tabularx}

\caption{
Performance before and after optimizing a pruning solution for the test-set images. Values are reported as mean $\pm$ standard deviation across 10 random initializations.
PLCC = Pearson correlation; SRCC = Spearman correlation.
\textit{Red. (\%)} indicates the percentage of feature maps removed.
}
\label{tab:accuracy}
\end{table}

\end{document}